\documentclass{clv3}

\usepackage{hyperref}
\usepackage{xcolor}
\usepackage{latexsym}
\usepackage{url}
\usepackage[T1]{fontenc}
\usepackage{graphicx}
\usepackage{color}
\usepackage{multirow}
\usepackage{enumitem}
\usepackage{mathrsfs}
\usepackage{booktabs}
\usepackage{pgfplots}
\usepackage{subcaption}
\usepackage{pifont}
\usepackage{amsmath,amsfonts}
\usepackage{arydshln}
\newcommand{\xmark}{\ding{55}}%
\newcommand{\cmark}{\ding{51}}%
\definecolor{darkblue}{rgb}{0, 0, 0.5}
\hypersetup{colorlinks=true,citecolor=darkblue, linkcolor=darkblue, urlcolor=darkblue}
\newcommand{\linestack}[1]{\def\arraystretch{0.7}\begin{tabular}[c]{@{}c@{}} #1 \end{tabular}}
\issue{1}{1}{2020}

\runningtitle{Machine Reading Comprehension: The Role of Contextualized Language Models and Beyond}

\runningauthor{Zhang et al. }

\begin{document}

\title{Machine Reading Comprehension: \\The Role of Contextualized Language Models and Beyond}

\author{Zhuosheng Zhang}
\affil{Shanghai Jiao Tong University \\ Department of Computer Science and Engineering \\ \tt zhangzs@sjtu.edu.cn}

\author{Hai Zhao}
\affil{Shanghai Jiao Tong University \\ Department of Computer Science and Engineering \\ \tt zhaohai@cs.sjtu.edu.cn}

\author{Rui Wang}
\affil{National Institute of Information and Communications Technology (NICT) \\ \tt wangrui@nict.go.jp}

\maketitle
\begin{abstract}
Machine reading comprehension (MRC) aims to teach machines to read and comprehend human languages, which is a long-standing goal of natural language processing (NLP). 
With the burst of deep neural networks and the evolution of contextualized language models (CLMs), the research of MRC has experienced two significant breakthroughs. MRC and CLM, as a phenomenon, have a great impact on the NLP community.
In this survey, we provide a comprehensive and comparative review on MRC covering overall research topics about 1) the origin and development of MRC and CLM, with particular focus on the role of CLMs; 2) the impact of MRC and CLM to the NLP community; 3) the definition, datasets, and evaluation of MRC; 4) general MRC architecture and technical methods in the view of two-stage Encoder-Decoder solving architecture from the insights of the cognitive process of humans; 5) previous highlights, emerging topics, and our empirical analysis, among which we especially focus on what works in different periods of MRC researches. 
We propose a full-view categorization and new taxonomies on these topics.
The primary views we have arrived at are that
1) MRC boosts the progress from language processing to understanding; 2) the rapid improvement of MRC systems greatly benefits from the development of CLMs; 3) the theme of MRC is gradually moving from shallow text matching to cognitive reasoning. 
\end{abstract}

\section{Introduction}
\begin{figure*}[htb]
    \centering
    \includegraphics[width=1.0\textwidth]{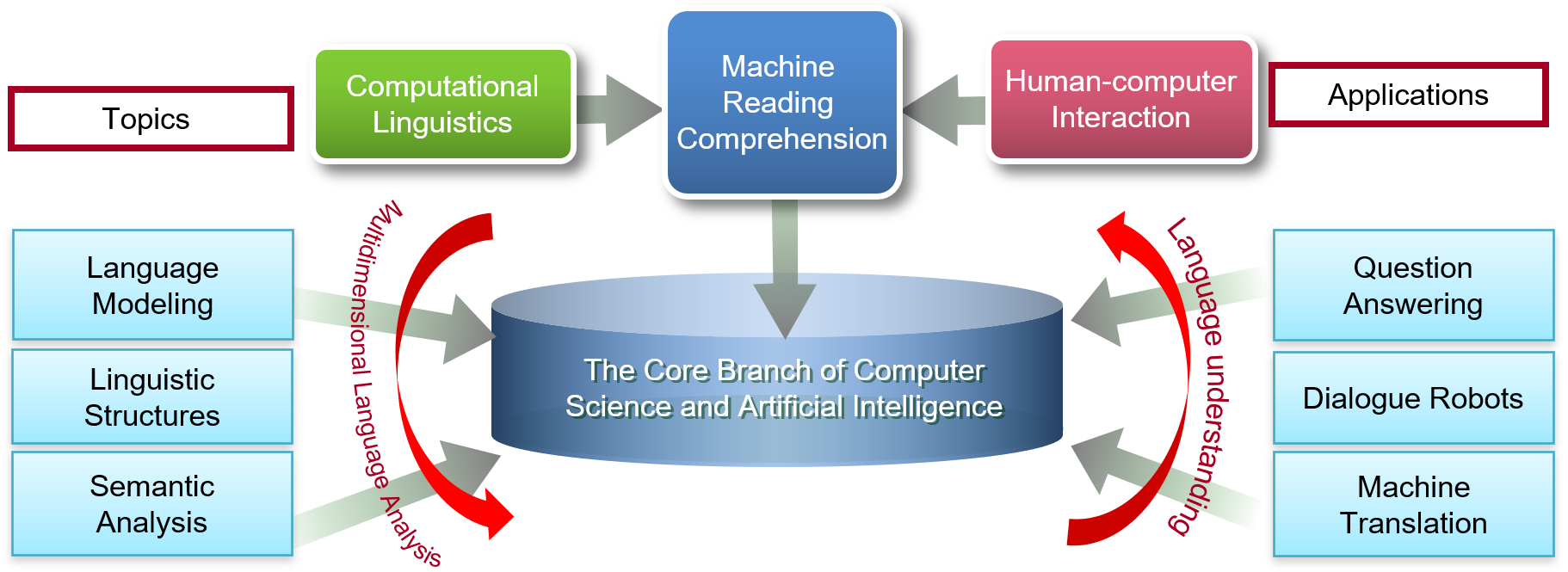}
    \caption{Overview of language processing and understanding.}
    \label{overview_nlp}
\end{figure*}

Natural language processing (NLP) tasks can be roughly divided into two categories: 1) fundamental NLP, including language modeling and representation, and linguistic structure and analysis, including morphological analysis, word segmentation, syntactic, semantic and discourse paring, etc.; 2) application NLP, including machine question answering, dialogue system, machine translation, and other language understanding and inference tasks. With the rapid development of NLP, natural language understanding (NLU) has aroused broad interests, and a series of NLU tasks have emerged. In the early days, NLU was regarded as the next stage of NLP. With more computation resources available, more complex networks become possible, and researchers are inspired to move forward to the frontier of human-level language understanding. Inevitably, machine reading comprehension (MRC) \cite{richardson2013mctest,hermann2015teaching,hill2015goldilocks,rajpurkar2016squad} as a new typical task has boomed in the field of NLU. Figure \ref{overview_nlp} overviews MRC in the background of language processing and understanding.

\begin{figure*}[htb]
    \includegraphics[width=0.8\textwidth]{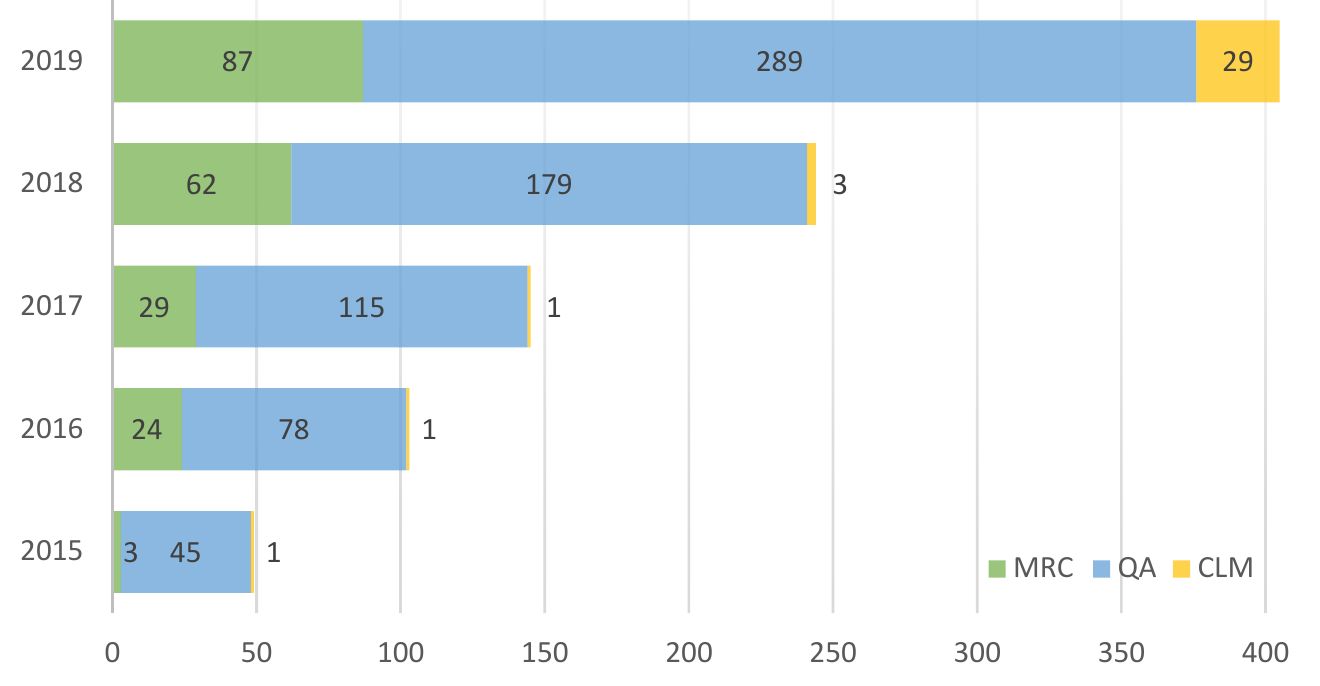}
    \caption{The number of papers concerning MRC, QA, and CLM collected from 2015 to 2019. The search terms are MRC: \{machine reading comprehension, machine comprehension, machine comprehend, mrc\}; QA: \{question answering, qa\}. Since MRC papers are often in the name of QA, we also present the QA papers for reference. MRC and QA papers are searched by keywords in paper titles on \url{https://arxiv.org}. CLM statistics are calculated based on the influential open-source repository: \url{https://github.com/thunlp/PLMpapers}.  }
    \label{fig:papers}
\end{figure*}

MRC is a long-standing goal of NLU that aims to teach a machine to read and comprehend textual data. It has significant application scenarios such as question answering and dialog systems \cite{choi2018quac,reddy2019coqa,zhang2018dua,zhu2018lingke,xu2020matinf}. The related MRC research can be traced back to the studies of story comprehension \cite{lehnert1977conceptual,cullingford1977controlling}. After decades of decline, MRC becomes a hot research topic recently and experiences rapid development. MRC has a critical impact on NLU and the broader NLP community. As one of the major and challenging problems of NLP concerned with comprehensive knowledge representation, semantic analysis, and reasoning, MRC stimulates great research interests in the last decade.
% the research interests of MRC reflects the core development of NLP, and promote the other NLP tasks. 
%  MRC is a challenging problem  including CLMs. 
The study of MRC has experienced two significant peaks, namely, 1) the burst of deep neural networks; 2) the evolution of contextualized language models (CLMs). Figure \ref{fig:papers} shows the research trend statistics of MRC and CLMs in the past five years.

Early MRC task was simplified as requiring systems to return a sentence that contains the right answer. The systems are based on rule-based heuristic methods, such as bag-of-words approaches \cite{hirschman1999deep}, and manually generated rules \cite{riloff2000rule,charniak2000reading}. With the introduction of deep neural networks and effective architecture like attention mechanisms in NLP \cite{bahdanau2014neural,hermann2015teaching}, the research interests of MRC boomed since around 2015 \cite{chen2016thorough,Nguyen2016MSMA,rajpurkar2016squad,trischler2017newsqa,dunn2017searchqa,he2018dureader,kovcisky2018narrativeqa,yang2018hotpotqa,reddy2019coqa,pan2019frustratingly}. The main topics were fine-grained text encoding and better passage and question interactions \cite{Seo2016Bidirectional,Yang2016Words,dhingra2017gated,Cui2017Attention,zhang2018effective}. 

CLMs lead to a new paradise of contextualized language representations --- using the whole sentence-level representation for language modeling as pre-training, and the context-dependent hidden states from the LM are used for downstream task-specific fine-tuning.
Deep pre-trained CLMs \cite{peters2018deep,devlin2018bert,yang2019xlnet,lan2019albert,dong2019unified,clark2019electra,joshi2020spanbert} greatly strengthened the capacity of language encoder, the benchmark results of MRC were boosted remarkably, which stimulated the progress towards more complex reading, comprehension, and reasoning systems \cite{welbl2018constructing,yang2018hotpotqa,ding2019cognitive}. As a result, the researches of MRC become closer to human cognition and real-world applications. On the other hand, more and more researchers are interested in analyzing and interpreting how the MRC models work, and investigating the \textit{real} ability beyond the datasets, such as performance in the adversarial attack \cite{jia2017adversarial,wallace2019trick}, as well as the benchmark capacity of MRC datasets \cite{sugawara2018makes,sugawara2019assessing,schlegel2020framework}. The common concern is the over-estimated ability of MRC systems, which shows to be still in a shallow comprehension stage drawn from superficial pattern-matching heuristics. Such assessments of models and datasets would be suggestive for next-stage studies of MRC methodologies.

MRC is a generic concept to probe for language understanding capabilities \cite{schlegel2020framework,gardner2019question}. In the early stage, MRC was regarded as the form of triple-style (passage, question, answer) question answering (QA) task, such as the cloze-style \cite{hermann2015teaching,hill2015goldilocks}, multiple-choice \cite{lai2017race,sun2019dream}, and span-QA \cite{rajpurkar2016squad,Rajpurkar2018Know}. In recent years, we witness that the concept of MRC has evolved to a broader scope, which caters to the theme of language understanding based interaction and reasoning, in the form of question answering, text generation, conversations, etc. Though MRC originally served as the form of question answering, it can be regarded as not only just the extension of QA but also a new concept used for studying the capacity of language understanding over some context that is close to cognitive science, instead of a single task itself. Regarding MRC as phenomenon, there is a new emerging interest showing that classic NLP tasks can be cast as span-QA MRC form, with modest performance gains than previous methodologies \cite{mccann2018natural,keskar2019unifying,li2019entity,li2019unified,keskar2019unifying,gao2019dialog,gao2020machine}.

Although it is clear that computation power substantially fuels the capacity of MRC systems in the long run, building simple, explainable, and practical models is equally essential for real-world applications. It is instructive to review the prominent highlights in the past. The generic nature, especially what works in the past and the inspirations of MRC to the NLP community, would be suggestive for future studies, which are the focus of discussions in this work. 

This work reviews MRC covering the scope of background, definition, influence, datasets, technical and benchmark success, empirical assessments, current trends, and future opportunities.
Our main contributions are summarized as follows:
\begin{itemize}
\item \textbf{Comprehensive review and in-depth discussions}. We conduct a comprehensive review of the origin and the development of MRC, with a special focus on the role of CLMs. We propose new taxonomies of the technical architecture of MRC, by formulating the MRC systems as two-stage solving architecture in the view of cognition psychology and provide a comprehensive discussion of research topics to gain insights. By investigating typical models and the trends of the main flagship datasets and leaderboards concerning different types of MRC, along with our empirical analysis, we provide observations of the advances of techniques in different stages of studies.
\item \textbf{Wide coverage on highlights and emerging topics}. MRC has experienced rapid development. We present a wide coverage of previous highlights and emerging topics, including casting traditional NLP tasks into MRC formation, multiple granularity feature modeling, structured knowledge injection, contextualized sentence representation, matching interaction, and data augmentation.
\item \textbf{Outlook on the future}. This work summaries the trends and discussions for future researches, including interpretability of datasets and models, decomposition of prerequisite skills, complex reasoning, large-scale comprehension, low-resource MRC, multimodal semantic grounding, and deeper but efficient model design.
\end{itemize}

The remainder of this survey is organized as follows:
first, we present the background, categorization, and derivatives of CLM and discuss the mutual influence between CLM and MRC in \S\ref{sec:role_of_CLM}; an overview of MRC including the impact to general NLP scope, formations, datasets, and evaluation metrics is given in \S\ref{sec:mrc_as_pheno}; then, we discuss the technical methods in the view of two-stage solving architecture, and summarize the major topics and challenges in \S\ref{sec:techniques}; next, our work goes deeper in \S\ref{sec:reallyworks} to discover what works in different stages of MRC, by reviewing the trends and highlights entailed in the typical MRC models. Our empirical analysis is also reported for the verification of simple and effective tactic optimizations based on the strong CLMs; finally, we discuss the trends and future opportunities in  \S\ref{sec:trends}, together with conclusions in  \S\ref{sec:conclu}; 

\section{The Role of Contextualized Language Model}\label{sec:role_of_CLM}

\subsection{From Language Model to Language Representation}
Language modeling is the foundation of deep learning methods for natural language processing. Learning word representations has been an active research area, and aroused great research interests for decades, including non-neural \cite{brown1992class,ando2005framework,blitzer2006domain} and neural methods \cite{mikolov2013distributed,pennington2014glove}. Regarding language modeling, the basic topic is $n$-gram language model (LM). An $n$-gram Language model is a probability distribution over word ($n$-gram) sequences, which can be regarded with a training objective of predicting unigram from ($n-1$)-gram. Neural networks use continuous and dense representation, or further embedding of words to make their predictions, which is effective for alleviating the curse of dimensionality -- as language models are trained on larger and larger texts, the number of unique words increases.

Compared with the word embeddings learned by Word2Vec \cite{mikolov2013distributed} or GloVe \cite{pennington2014glove}, sentence is the least unit that delivers complete meaning as human uses language. Deep learning for NLP quickly found it is a frequent requirement on using a network component encoding a sentence input so that we have the \textit{Encoder} for encoding the complete sentence-level context.
The encoder can be the traditional RNN, CNN, or the latest Transformer-based architectures, such as ELMo \cite{peters2018deep}, GPT$_{v1}$ \cite{radford2018improving}, BERT \cite{devlin2018bert}, XLNet \cite{yang2019xlnet},  RoBERTa \cite{liu2019roberta}, ALBERT \cite{lan2019albert}, and ELECTRA \cite{clark2019electra}, for capturing the contextualized sentence-level language representations.\footnote{This is a non-exhaustive list of important CLMs introduced recently. In this work, our discussions are mainly based on these typical CLMs, which are highly related to MRC researches, and most of the other models can be regarded as derivatives.}
These encoders differ from sliding window input (e.g., that used in Word2Vec) that they cover a full sentence instead of any fixed length sentence segment used by the sliding window. Such difference especially matters when we have to handle passages in MRC tasks, where the passage always consists of a lot of sentences. When the model faces passages, the sentence, instead of word, is the basic unit of a passage.  In other words, MRC, as well as other application tasks of NLP, needs a sentence-level encoder, to represent sentences into embeddings, so as to capture the deep and contextualized sentence-level information. 

\begin{table}[t]
\caption{Comparison of language representation.}
  \setlength{\tabcolsep}{3.5pt}
\begin{tabular}{lllll}
\toprule
Model & Repr. form &  Context & Training object & Usage\\
\midrule
$n$-gram LM & One-hot & Sliding widow & $n$-gram LM (MLE) & Lookup\\
Word2vec/GloVe & Embedding & Sliding widow & $n$-gram LM (MLE) & Lookup\\
Contextualized LM & Embedding & Sentence & $n$-gram LM (MLE), +ext & Fine-tune\\
\end{tabular}
\label{tab:language_representation}
\end{table}

An encoder model can be trained in a style of $n$-gram language model so that there comes the language representation, which includes four elements: 1) representation form; 2) context; 3) training object (e.g., $n$-gram language model); 4) usage. For contextualized language representation, the representation for each word depends on the entire context in which it is used, which is dynamic embedding. Table \ref{tab:language_representation} presents a comparison of the three main language representation approaches.

\subsection{CLM as Phenomenon}
\subsubsection{Revisiting the Definition}
First, we would like to revisit the definitions of the recent contextualized encoders. For the representative models, ELMo is called \textit{Deep contextualized word representations}, and BERT \textit{Pre-training of deep bidirectional transformers for language understanding}. With the follow-up research goes on, there are studies that call those models as pre-trained (language) models \cite{sanh2019distilbert,goldberg2019assessing}. We argue that such a definition is reasonable but not accurate enough. The focus of these models are supposed to be \textit{contextualized} (as that show in the name of ELMo), in terms of the evolution of language representation architectures, and the actual usages of these models nowadays. As a consensus of limited computing resources, the common practice is to fine-tune the model using task-specific data after the public pre-trained sources, so that pre-training is neither the necessary nor the core element. As shown in Table \ref{tab:language_representation}, the training objectives are derived from $n$-gram language models. Therefore, we argue that pre-training and fine-tuning are just the manners we use the models. The essence is the deep contextualized representation from language models; thus, we call these pre-trained models \textbf{contextualized language models, CLMs)} in this paper.

\subsubsection{Evolution of CLM Training Objectives}
In this part, we abstract the inherent relationship of $n$-gram language model and the subsequent contextualized LM techniques. Then, we elaborate the evolution of the typical CLMs considering the salient role of the training objectives.

Regarding the training of language models, the standard and common practice is using the $n$-gram language modeling. It is also the core training objective in CLMs. An $n$-gram Language model yields a probability distribution over text ($n$-gram) sequences, which is a classic maximum likelihood estimation (MLE) problem. The language modeling is also known as \textbf{autoregressive} (AR) scheme.
\begin{figure*}[htb]
\centering
    \includegraphics[width=0.5\textwidth]{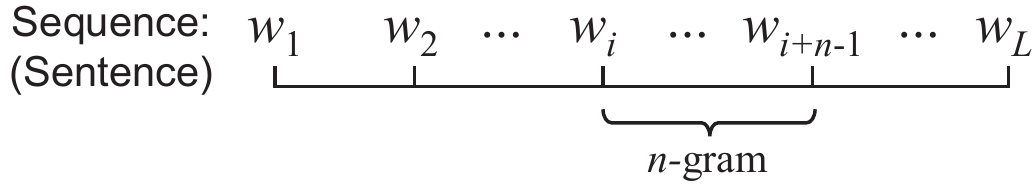}
    \caption{Example of $n$-grams.}
    \label{fig:ngram}
\end{figure*}

Specifically, given a sequence of $n$ items $\textbf{w}=w_{i:i+n-1}$ from a text (Figure \ref{fig:ngram}), the probability of the sequence is measured as
\begin{align}
p(\textbf{w}) = p(w_i\mid w_{i:i+n-2}),
\end{align}
where $p(w_i|w_{i:i+n-2})$ denotes the conditional probability of $p(w_i)$ in the sequence, which can be estimated by the context representation over $w_{i:i+n-2}$. 
% \begin{align}
% p(w_k|\textbf{w}_{1:k-1}) = \mathcal{G}(\textbf{w}_{1:k-1}),
% \end{align}
The LM training is performed by maximizing the likelihood:
\begin{align}
\max_{\theta} \sum_{\textbf{w}} \log p_{\theta}(\textbf{w}),
\label{eq:ngram}
\end{align}
where $\theta$ denotes the model parameter.

In practice, $n$-gram models have been shown to be extremely effective in modeling language data, which is a core component in modern language applications. The early contextualized representation is obtained by static word embedding and a network encoder. For example, CBOW and Skip-gram
\cite{mikolov2013distributed} either predicts the word using context or predict context by word, where the $n$-gram context is provided by a fixed sliding window. The trained model parameters are output as a word embedding matrix (also known as a lookup table), which contains the context-independent representations for each word in a vocabulary.
The vectors are then used in a low-level layer (i.e., embedding layer) of neural network, and an encoder, such as RNN is further used to obtain the contextualized representation for an input sentence. 

For recent LM-derived \textbf{contextualized} presentations \cite{peters2018deep,devlin2018bert,yang2019xlnet}, the central point of the subsequent optimizations are concerning the context. They are trained with much larger $n$-grams that cover a full sentence where $n$ is extended to the sentence length --- \textbf{when $n$ expands to the maximum, the conditional context thus corresponds to the whole sequence}. The word representations are the function of the entire sentence, instead of the static vectors over a pre-defined lookup table. The corresponding functional model is regarded as a contextualized language model. Such a contextualized model can be directly used to produce context-sensitive sentence-level representations for task-specific fine-tuning. Table \ref{tab:comparison-lms} shows the comparisons of CLMs.

\begin{figure*}
    \includegraphics[width=1.0\textwidth]{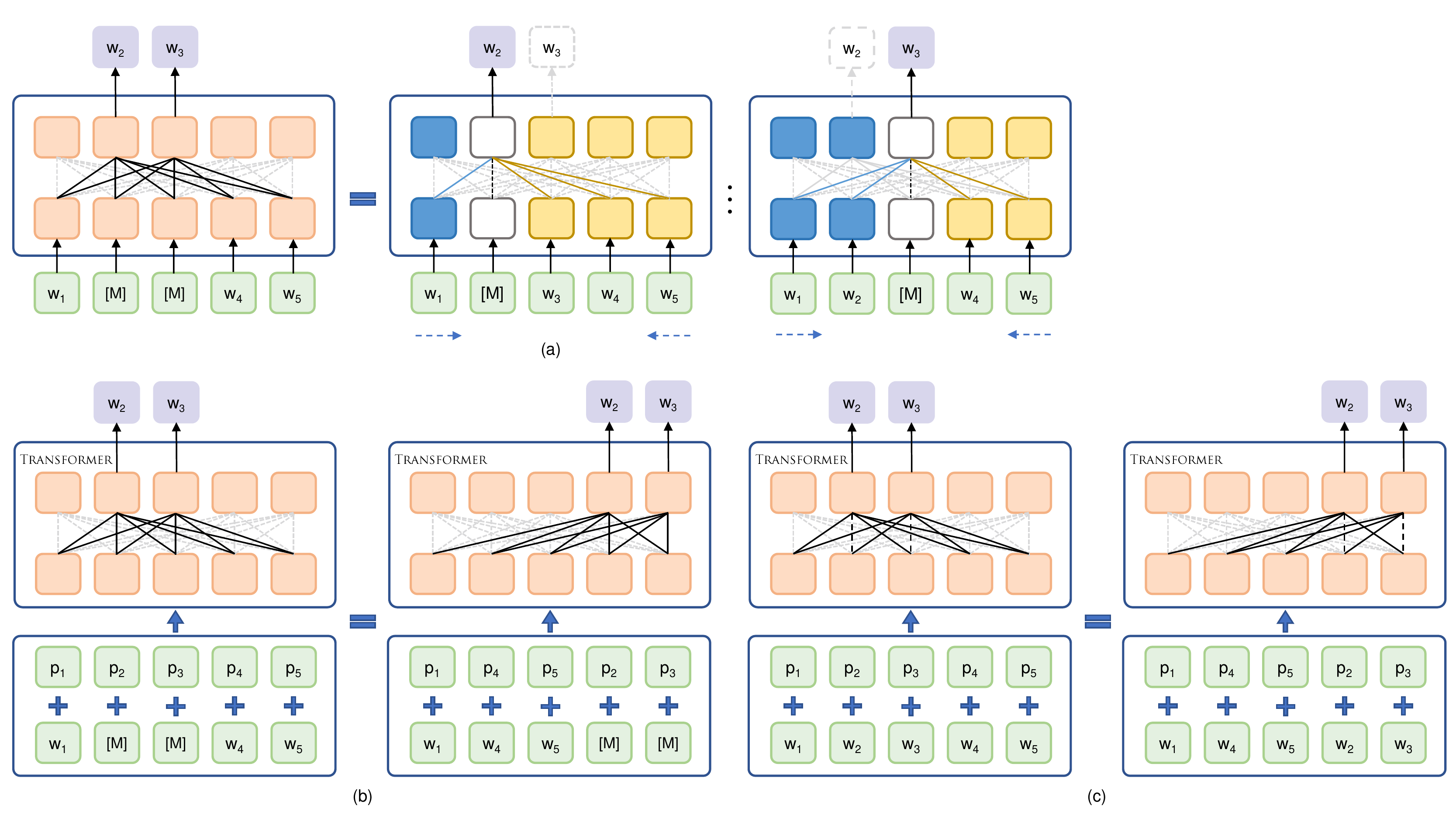}
    \caption{The possible transformation of MLM and PLM, where $w_i$ and $p_i$ represent token and position embeddings. $[M]$ is the special mask token used in MLM. The left side of MLM (a) can be seen as bidirectional AR streams (in \textcolor[RGB]{91,155,213}{blue} and \textcolor[RGB]{255,230,153}{yellow}, respectively) at the right side. For MLM (b) and PLM (c), the left sides are in original order, and the right sides are in permuted order, which are regarded as a unified view.}
    \label{Permuted_view}
\end{figure*}

For an input sentence $\mathbf{s}=w_{1:L}$, we extend the objective of $n$-gram LM in the context of length $L$ from Equation~(\ref{eq:ngram}): 
\begin{align}
\sum_{k=c+1}^{L} \log p_{\theta}(w_k\mid w_{1:k-1}),
\label{eq:ngram_uni}
\end{align}
 where $c$ is the cutting point that separate the sequence into a non-target conditional subsequence $k \leq c$ and a target subsequence $k > c$. It can be further written in a bidirectional form:
\begin{align}
\sum_{k=c+1}^{L} (\log p_{\theta}(w_k\mid w_{1:k-1}) + \log p_{\theta}(w_k\mid w_{k+1:L})),
\label{eq:bilm}
\end{align}
which corresponds to the bidirectional LM used in ELMo \cite{peters2018deep}. The bidirectional modeling of ELMo is achieved by the concatenation of independently trained forward and backward LSTMs.  

To allow simultaneous bidirectional (or non-directional) training, BERT \cite{devlin2018bert} adopted Transformer to process the whole input at once, and proposed Masked LM (MLM) to take advantage of both the left and right contexts. Some tokens in a sentence are randomly replaced with a special mask symbol with a small probability. Then, the model is trained to predict the masked token based on the context. 
MLM can be seen as a variant of $n$-gram LM (Figure \ref{Permuted_view}(a)) to a certain extent --- bidirectional autoregressive $n$-gram LM.\footnote{In a general view, the idea of MLM can also be derived from CBOW, which is to predict word according to the conditional $n$-gram surrounding context.}
Let $\mathcal{D}$ denote the set of masked positions using the mask symbol $[M]$. We have $w_\mathcal{D}$ as the set of masked tokens, and $\mathbf{s}'$ as the masked sentence. As the example shown in the left part of Figure~\ref{Permuted_view}(b), $\mathcal{D} = \{2, 3\}$, $w_{\mathcal{D}} = \{w_2, w_3\}$ and $\mathbf{s}' = \{w_1, [M], w_4, [M], w_5\}$. The objective of MLM is to maximize the following objective:
	\begin{equation}
	\sum\limits_{k \in \mathcal{D}} \log p_{\theta}(w_k\mid\mathbf{s}')
	\label{eq_mlm_org} 
	\end{equation}

Compared with Equation~(\ref{eq:bilm}), it is easy to find that the prediction is based on the whole context in Equation~(\ref{eq_mlm_org}) instead of only one direction for each estimation, which indicates the major difference of BERT and ELMo. However, the essential problem in BERT is that the mask symbols are never seen at fine-tuning, which faces a mismatch between pre-training and fine-tuning. 

\begin{table}
\caption{Comparison of CLMs. NSP: next sentence prediction \cite{devlin2018bert}. SOP: sentence order prediction \cite{lan2019albert}. RTD: replaced token detection \cite{clark2019electra}.}
  \setlength{\tabcolsep}{2.8pt}
\begin{tabular}{llllll}
\toprule
Model & Loss &  $2^{nd}$ Loss & Direction & Encoder arch. & Input \\
\midrule
ELMo & $n$-gram LM & - & Bi & RNN & Char \\
GPT$_{v1}$ & $n$-gram LM & - & Uni & Transformer & Subword \\
BERT & Masked LM & NSP & Bi & Transformer & Subword \\
\quad RoBERTa & Masked LM & - & Bi & Transformer & Subword\\
\quad ALBERT & Masked LM & SOP & Bi & Transformer & Subword \\
XLNet & Permu. $n$-gram LM & - & Bi & Transformer-XL & Subword \\
ELECTRA & Masked LM & RTD & Bi & GAN & Subword\\
\end{tabular}
\label{tab:comparison-lms}
\end{table}

To alleviate the issue, XLNet \cite{yang2019xlnet} utilized permutation LM (PLM) to maximize the expected log-likelihood of all possible permutations of the factorization order, which is the AR LM objective.\footnote{In contrast, the language modeling method in BERT is called denoising \textbf{autoencoding} \cite{yang2019xlnet} (AE). AE can be seen as the natural combination of AR loss and a certain neural network.}  
For the input sentence $\textbf{s}=w_{1:L}$, we have $\mathcal{Z}_L$ as the permutations of set $\{1, 2, \cdots, L\}$. For a permutation $z \in \mathcal{Z}_L$,  we split $z$ into a non-target conditional subsequence $z \leq c$ and a target subsequence $z > c$, where $c$ is the cutting point.
The objective is to maximize the log-likelihood of the target tokens conditioned on the non-target tokens:
	\begin{equation}
	 \mathbb{E}_{z \in \mathcal{Z}_L} \sum\limits_{k=c+1}^{L} \log  p_{\theta}(w_{z_k}\mid w_{z_{1:k-1}}).
	\label{eq2_plm}
	\end{equation}

The key of both MLM and PLM is predicting word(s) according to a certain context derived from $n$-grams, which can be modeled in a unified view \cite{song2020mpnet}. In detail, under the hypothesis of word order insensitivity, MLM can be directly unified as PLM when the input sentence is permutable (with insensitive word orders), as shown in Figure \ref{Permuted_view}(b-c). It can be satisfied thanks to the nature of the Transformer-based models, such as BERT and XLNet. Transformer takes tokens and their positions in a sentence as inputs, and it is not sensitive to the absolute input order of these tokens. 
Therefore, the objective of MLM can be also written as the permutation form,
\begin{equation}
	\mathbb{E}_{z \in \mathcal{Z}_L} \sum\limits_{k=c+1}^{L} \log  p_{\theta}(w_{z_k}\mid w_{z_{1:c}}, M_{z_{k:L}}),
	\label{eq3_mlm_unify}
	\end{equation}
	where $M_{z_{k:L}}$ denote the special mask tokens $[M]$ in positions $z_{k:L}$. 

From Equations~(\ref{eq:ngram_uni}), (\ref{eq2_plm}), and (\ref{eq3_mlm_unify}), we see that MLM and PLM share similar  formulations with the $n$-gram LM with slight difference in the conditional context part in $p(\textbf{s})$: MLM conditions on $w_{z_{1:c}}$ and $M_{k:L}$, and PLM conditions on $w_{z_{1:k-1}}$. \textbf{Both MLM and PLM can be explained by the $n$-gram LM, and even unified into a general formation}. With similar inspiration, MPNet \cite{song2020mpnet} combined the Masked LM and Premuted LM for taking both of the advantages. 

\subsubsection{Architectures of CLMs}
\begin{figure*}
    \includegraphics[width=1.0\textwidth]{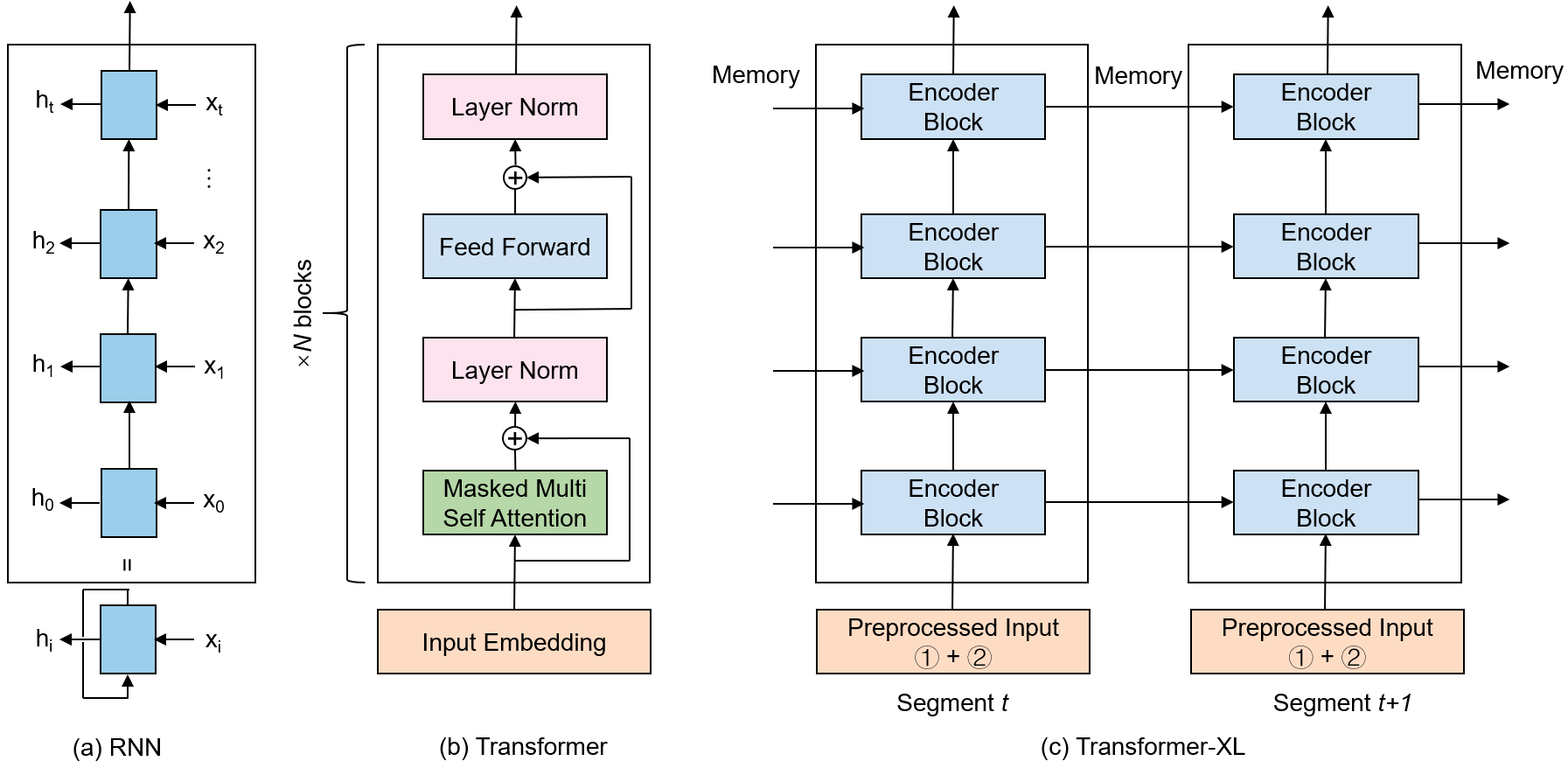}
    \caption{RNN, Transformer, and Transformer-XL encoder architectures for CLMs.}
    \label{fig:encoder_ark}
\end{figure*}

So far, there are mainly three leading architectures for language modeling,\footnote{Actually, CNN also turns out well-performed feature extractor for some NLP tasks like text classification, but RNN is more widely used for MRC, even most NLP tasks; thus we omit the description of CNNs and focus on RNNs as the example for traditional encoders.} RNN, Transformer, and Transformer-XL. Figure \ref{fig:encoder_ark} depicts the three encoder architectures.

\paragraph{RNN} RNN and its derivatives are popular approaches for language encoding and modeling. The widely-used variants are GRU \cite{cho2014learning} and LSTM \cite{hochreiter1997long}. RNN models process the input tokens (commonly words or characters) one by one to capture the contextual representations between them. However, the processing speed of RNNs is slow, and the ability to learn long-term dependencies is still limited due to vanishing gradients. 

\paragraph{Transformer} To alleviate the above issues of RNNs, Transformer was proposed, which employs \textit{multi-head self-attention} \cite{NIPS2017_7181} modules receive a segment of tokens (i.e., subwords) and the corresponding position embedding as input to learn the direct connections of the sequence at once, instead of processing tokens one by one.

\paragraph{Transformer-XL} Though both RNN and Transformer architectures have reached impressive achievements, their main limitation is capturing long-range dependencies. Transformer-XL \cite{dai2019transformer} combines the advantages of RNN and Transformer, which uses the self-attention modules on each segment of input data and a recurrent mechanism to learn dependencies between consecutive segments. In detail, two new techniques are proposed:
\begin{enumerate}
    \item \textbf{Segment-level Recurrence}. The recurrence mechanism is proposed to model long-term dependencies by using information from previous segments. During training, the representations computed for the previous segment are fixed and cached to be reused as an extended context when the model processes the next new segment. This recurrence mechanism is also effective in resolving the context fragmentation issue, providing necessary context for tokens in the front of a new segment.
    \item \textbf{Relative Positional Encoding}.  The original positional encoding deals with each segment separately. As a result, the tokens from different segments have the same positional encoding. The new relative positional encoding is designed as part of each attention module, as opposed to the encoding position only before the first layer. It is based on the relative distance between tokens, instead of their absolute position.
\end{enumerate}

\begin{figure*}
    \includegraphics[width=1.0\textwidth]{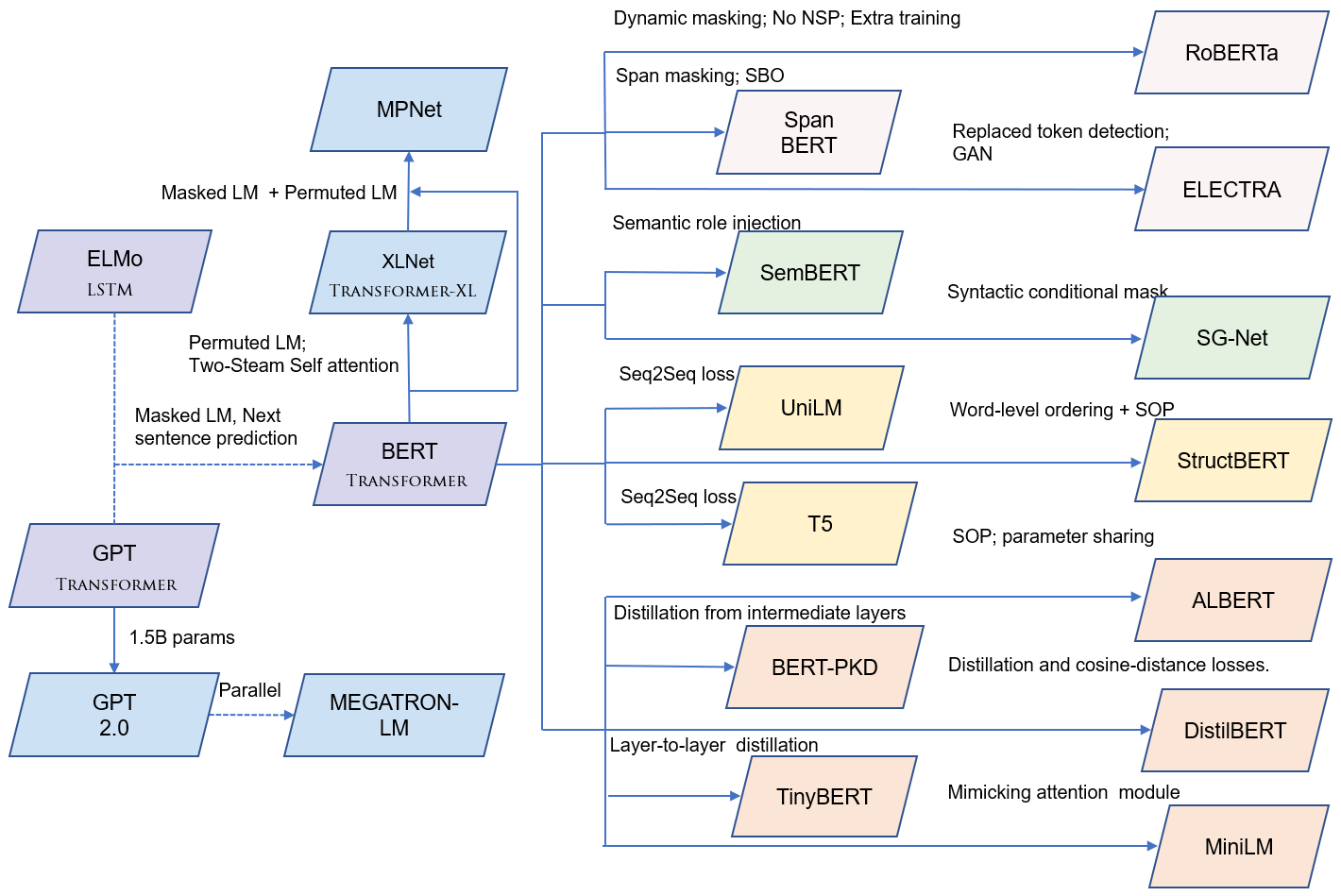}
    \caption{Derivative of CLMs. The main features are noted above the arrow. Solid and dotted arrows indicate the direct and implicit inheritance.}
    \label{fig:encoder_comp}
\end{figure*}

\subsubsection{Derivative of CLMs}
Pre-training and fine-tuning have become a new paradigm of NLP, and the major theme is to build a strong encoder. Based on the inspirations of impressive models like ELMo and BERT, a wide range of CLMs derivatives have been proposed. In this part, we discuss various major variants concerning MRC tasks. Table \ref{tab:perfom_clms} shows the performance comparison of the CLM derivatives. The advances behind these models are in four main topics:
\paragraph{Masking Strategy} The original masking of BERT is based on subword, which would be insufficient for capturing global information using the local subword signals. 
% BERT WWM improved the masking mechanism via whold-word dimension for word-level prediction. 
% ERNIE \cite{sun2019ernie} masked the word, phrase, and entities through prior knowledge from extra sources. ERNIE-THU \cite{zhang2019ernie} randomly masked some of the named entity alignments in the input text, requiring the model to select appropriate entities from knowledge graphs to complete the alignments. In contrast, 
SpanBERT \cite{joshi2020spanbert} proposed a random span masking strategy based on geometric distribution, indicating that the proposed masking sometimes works even better than masking linguistically-coherent spans. To avoid using the same mask for each training instance in every epoch, RoBERTa \cite{liu2019roberta} used dynamic masking to generate the masking pattern every time feeding a sequence to the model, indicating that dynamic masking would be crucial for pre-training a great many steps or with large-scale datasets. ELECTRA \cite{clark2019electra} improved the efficiency of masking by adopting a replaced token detection objective. 
\paragraph{Knowledge Injection}
Extra knowledge can be easily incorporated into CLMs by both embedding fusion and masking. SemBERT \cite{zhang2020semantics} indicated that fusing semantic role label embedding and word embedding can yield better semantic-level language representation, showing that salient word-level high-level tag features can be well integrated with subword-level token representations. SG-Net \cite{zhang2019sg} presented a dependency-of-interest masking strategy to use syntax information as a constraint for better linguistics inspired representation.

\begin{table}
\caption{Performance of CLM derivatives. F1 scores for SQuAD1.1 and SQuAD2.0, accuracy for RACE. * indicates results that depend on additional data augmentation. $\dagger$ indicate the result is from \citet{yang2019xlnet} as it was not reported in the original paper \cite{devlin2018bert}. The BERT$_{base}$ result for SQuAD2.0 is from \citet{wang2020minilm}. The \textit{italic} numers are baselines for calculating the D-values $\uparrow$.}
  \setlength{\tabcolsep}{3pt}
\begin{tabular}{lrrrrrrrrrrr}
\toprule
\multirow{2}{*}{Method}  & \multicolumn{4}{c}{SQuAD1.1} & \multicolumn{4}{c}{SQuAD2.0} & \multicolumn{2}{c}{RACE} & \\  & Dev &  $\uparrow$ Dev & Test &  $\uparrow$ Test & Dev &  $\uparrow$ Dev & Test &  $\uparrow$ Test & Acc & $\uparrow$ Acc\\
\midrule
ELMo & \textit{85.6}  & - & \textit{85.8}  & - & - &  & -  & & -  & - \\
GPT$_{v1}$ &  -  & - & -  & - & -  & - & -  & - &   \textit{59.0}  & -\\
BERT$_{base}$ &   88.5  & 2.9 & -   & - & \textit{76.8} &  & -  &  &  65.3 & 6.3\\
\quad BERT-PKD & 85.3  & -0.3 & -  & - & 69.8  & -7.0 & - & - &  60.3  & 1.3 \\
\quad DistilBERT  & 86.2  & 0.6 &  -  & - & 69.5  & -7.3 & -  & -\\
\quad  TinyBERT & 87.5  & 1.9 & -  & - & 73.4  & -3.4 & -  & - & -  & -\\
\quad MiniLM & - & - & - & - & 76.4  & -0.4& - & - & - &- \\
\quad  Q-BERT & 88.4 & 2.8 &  - & - & - & - & -& - & -&- \\
BERT$_{large}$ & 91.1*  & 5.5 & 91.8*  & 6 & 81.9  & 5.1& 83.0 & - & 72.0$\dagger$  & -\\
% RoBERTa & Wikipedia + BooksCorpus & 16GB  &  355M & 93.6 & 87.3  & \\
\quad SemBERT$_{large}$ &  -  & - & -  & - & 83.6  & 6.8 & 85.2   & 2.2 & -  & - \\
\quad SG-Net &  -  & - & -   & - & 88.3  & 11.5 & 87.9 & 4.9 & 74.2  &  15.2 \\
\quad SpanBERT$_{large}$ & -  & - & 94.6  & 8.8 & -  & - & 88.7  & 5.7& -  & - \\
% \quad MPNet$_{base}$ & - & 92.5 &  - & 85.6 & \\
\quad StructBERT$_{large}$ & 92.0  & 6.4& - & - & -  & -& -  & -& -  & - \\
\quad RoBERTa$_{large}$ & 94.6  &  9.0& -  & - & 89.4  & 12.6 & 89.8  & 6.8 & 83.2  & 24.2\\
\quad ALBERT$_{xxlarge}$ & 94.8  & 9.2& -  & - & 90.2  &13.4 & 90.9  & 7.9 & 86.5  & 27.5\\
XLNet$_{large}$  & 94.5  & 8.9 & 95.1*  & 9.3& 88.8  & 12& 89.1*  & 6.1 &  81.8  & 22.8\\
UniLM & -   & - & -  & - & 83.4  & 6.6& -  & - & -  & -\\
ELECTRA$_{large}$  & 94.9  & 9.3& -  & -& 90.6 & 13.8& 91.4  & 8.4& -  & -\\
Megatron-LM$_{3.9B}$ &  95.5 & 9.9 &  -  & - & 91.2  & 14.4 &  -  & -  & 89.5  & 30.5\\
T5$_{11B}$ & 95.6  & 10.0 & -  & -& -  & - &  -  & - & -  & - \\

\end{tabular}
\label{tab:perfom_clms}
\end{table}

\paragraph{Training Objective} Besides the core MLE losses that used in language models, some extra objectives were investigated for better adapting target tasks. BERT \cite{devlin2018bert} adopted the next sentence prediction (NSP) loss, which matches the paired form in NLI tasks. To better model inter-sentence coherence, ALBERT \cite{lan2019albert} replaced NSP loss with a sentence order prediction (SOP) loss. StuctBERT \cite{Wang2020StructBERT} further leveraged word-level ordering and sentence-level ordering as structural objectives in pre-training. SpanBERT \cite{joshi2020spanbert} used span boundary objective (SBO), which requires the model to predict masked spans based on span boundaries, to integrate structure information into pre-training. UniLM \cite{dong2019unified} extended the mask prediction task with three types of language modeling tasks: unidirectional,  bidirectional, and sequence-to-sequence (Seq2Seq) prediction. The Seq2Seq MLM was also adopted as the objective in T5 \cite{raffel2019exploring}, which employed a unified Text-to-Text Transformer for general-purpose language modeling. ELECTRA \citet{clark2019electra} proposed new pre-training task --- replaced token detection (RTD) and a generator-discriminator model was designed accordingly. The generator is trained to perform MLM, and then the discriminator predicts whether each token in the corrupted input was replaced by a generator sample or not. 

\paragraph{Model Optimization}
RoBERTa \cite{liu2019roberta} found that the model performance can be substantially improved by 1) training the model longer, with bigger batches over more data can; 2) removing the next sentence prediction objective; 3) training on longer sequences; 4) dynamic masking on the training data.  Megatron \cite{shoeybi2019megatron} presented an intra-layer model-parallelism approach that can support efficiently training very large Transformer models. 

To obtain light-weight yet powerful models for real-world use, model compression is an effective solution. ALBERT \cite{lan2019albert} used cross-layer parameter sharing and factorized embedding parameterization to reduce the model parameters. Knowledge distillation (KD) also aroused hot interests. BERT-PKD proposed a patient KD mechanism that learns from multiple intermediate layers of the teacher model for incremental knowledge extraction. DistilBERT \cite{sanh2019distilbert} leveraged a knowledge distillation mechanism during the pre-training phase, which introduced a triple loss combining language modeling, distillation, and cosine-distance losses. TinyBERT \cite{jiao2019tinybert} adopted layer-to-layer distillation with embedding outputs, hidden states, and self-attention distributions. MiniLM \cite{wang2020minilm} performed the distillation on self-attention distributions and value relation of the teacher’s last Transformer layer to guide student model training. Moreover, quantization is another optimization technique by compressing parameter precision. Q-BERT \cite{shen2019q} applied a Hessian based mix-precision method to compress the mode with minimum loss in accuracy and more efficient inference.

\begin{table}[t]
\caption{The initial applications of CLMs. The concerned NLU task can also be regarded as a special case of MRC as discussed in \S\ref{sec:mrc_as_pheno}.}
  \setlength{\tabcolsep}{5.8pt}
\begin{tabular}{lcccccp{2.6cm}}
\toprule
& \multicolumn{2}{c}{NLU} & \multicolumn{3}{c}{MRC} &  \\
& SNLI & GLUE &  SQuAD1.1 & SQuAD2.0 & RACE  \\
\midrule
ELMo & \cmark  & \xmark & \cmark & \xmark & \xmark \\
GPT$_{v1}$ & \cmark & \cmark   & \xmark & \xmark & \cmark   \\
BERT & \xmark & \cmark  & \cmark & \cmark & \xmark  \\
\quad RoBERTa & \xmark & \cmark  & \cmark & \cmark & \cmark   \\
\quad ALBERT & \xmark & \cmark  & \cmark & \cmark & \cmark   \\
XLNet  & \xmark & \cmark  & \cmark & \cmark & \cmark   \\
ELECTRA  & \xmark & \cmark  & \cmark & \cmark & \xmark  \\
\end{tabular}
\label{tab:task-lms}
\end{table}

\subsection{Correlations Between MRC and CLM}
In the view of practice, MRC and CLM are complementary to each other. MRC is a challenging problem concerned with comprehensive knowledge representation, semantic analysis, and reasoning, which arouses great research interests and stimulates the development of wide ranges of advanced models, including CLMs. As shown in Table \ref{tab:task-lms}, MRC also serves as an appropriate testbed for language representation, which is the focus of CLMs. On the other hand, the progress of CLM greatly promotes MRC tasks, achieving impressive gains of model performance. With such an indispensable association, human-parity performance has been first achieved and frequently reported after the release of CLMs.

\section{MRC as Phenomenon}\label{sec:mrc_as_pheno}
\subsection{Classic NLP Meets MRC}
MRC has great inspirations to the NLP tasks. Most NLP tasks can benefit from the new task formation as MRC. The advantage may lie within both sides of 1) strong capacity of MRC-style models, e.g., keeping the pair-wise training mode like the pre-training of CLMs and better-contextualized modeling like multi-turn question answering \cite{li2019entity}; 2) unifying different tasks as MRC formation, and taking advantage of multi-tasking to share and transfer knowledge.
% \paragraph{Modeling traditional tasks as MRC paradigm}

Traditional NLP tasks can be cast as QA-formed reading comprehension over a context, including question answering, machine translation, summarization, natural language inference, sentiment analysis, semantic role labeling, zero-shot relation extraction, goal-oriented dialogue, semantic parsing, and commonsense pronoun resolution \cite{mccann2018natural}. The span extraction task formation of MRC also leads to superior or comparable performance for standard text classification and regression tasks, including those in GLUE benchmarks \cite{keskar2019unifying}, and entity and relation extraction tasks \cite{li2019entity,li2019unified,keskar2019unifying}. As MRC aims to evaluate how well machine models can understand human language, the goal is actually similar to the task of Dialogue State Tracking (DST). There are recent studies that formulate the DST task into MRC form by specially designing a question for each slot in the dialogue state, and propose MRC models for dialogue state tracking \cite{gao2019dialog,gao2020machine}. 
% \paragraph{Automatically generating large datasets} 

\subsection{MRC Goes Beyond QA}
In most NLP/CL papers, MRC is usually organized as a question answering task with respect to a given reference text (e.g., a passage). As discussed in \citet{chen2018neural}, there is a close relationship between MRC and QA. (Shallow) reading comprehension can be regarded as an instance of question answering, but they emphasize different final targets. We believe that the general MRC is a concept to probe for language understanding capabilities, which is very close to the definition of NLU. In contrast, QA is a format \cite{gardner2019question}, which is supposed to be the actual way to check how the machine comprehends the text. The rationale is the difficulty to measure the primary objective of MRC --- evaluating the degree of machine comprehension of human languages. To this end, QA is a fairly simple and effective format. MRC also goes beyond the traditional QA, such as factoid QA or knowledge base QA \cite{dong2015question} by reference to open texts, aiming at avoiding efforts on pre-engineering and retrieving facts from a structured manual-crafted knowledge corpus.

Therefore, though MRC tasks employ the form of question answering, it can be regarded as not only just the extension or variant of QA but also a new concept concerning studying the capacity of language understanding over some context. Reading comprehension is an old term to measure the knowledge accrued through reading. When it comes to machines, it concerns that machine is trained to read unstructured natural language texts, such as a book or a news article, comprehend and absorb the knowledge without the need of human curation. 

To some extent, traditional language understanding and inference tasks, such as textual entailment (TE), can be regarded as a type of MRC in theory as well. The common goal is to give a prediction after reading and comprehending the input texts; thus the NLI and standard MRC tasks are often evaluated together for assessing model's language understanding capacity \cite{peters2018deep,radford2018improving,Zhang2018subword,zhang2020SemBERT}. Besides, their forms can be converted to each other. MRC can be formed as NLI format \cite{zhang2019machine}, and NLI can also be regarded as multi-choice MRC (\textit{entailment}, \textit{neutral}, or \textit{contradictory}).

\begin{table}[htb]
	\caption{Examples of typical MRC forms.}
	\footnotesize
	\begin{tabular}{p{2cm}p{11cm}}
		\toprule
		Cloze-style  & from CNN \cite{hermann2015teaching} \\
		\midrule
		Context & \textit{( @entity0 ) -- a bus carrying members of a @entity5 unit overturned at an @entity7 military base sunday , leaving 23 @entity8 injured , four of them critically , the military said in a news release . a bus overturned sunday in @entity7 , injuring 23 @entity8 , the military said . the passengers , members of @entity13 , @entity14 , @entity15 , had been taking part in a training exercise at @entity19 , an @entity21 post outside @entity22 , @entity7 . they were departing the range at 9:20 a.m. when the accident occurred . the unit is made up of reservists from @entity27 , @entity28 , and @entity29 , @entity7 . the injured were from @entity30 and @entity31 out of @entity29 , a @entity32 suburb . by mid-afternoon , 11 of the injured had been released to their unit from the hospital . pictures of the wreck were provided to the news media by the military . @entity22 is about 175 miles south of @entity32 . e-mail to a friend} \\
		Question & \textit{bus carrying @entity5 unit overturned at \underline{\hbox to 8mm{}} military base} \\
		Answer & \textit{@entity7}  \\
		\midrule
		Multi-choice & from RACE \cite{lai2017race} \\
		\midrule
		Context & \textit{Runners in a relay race pass a stick in one direction. However, merchants passed silk, gold, fruit, and glass along the Silk Road in more than one direction. They earned their living by traveling the famous Silk Road.
			The Silk Road was not a simple trading network. It passed through thousands of cities and towns. It started from eastern China, across Central Asia and the Middle East, and ended in the Mediterranean Sea. It was used from about 200 B, C, to about A, D, 1300, when sea travel offered new routes, It was sometimes called the world’s longest highway. However, the Silk Road was made up of many routes, not one smooth path. They passed through what are now 18 countries. The routes crossed mountains and deserts and had many dangers of hot sun, deep snow, and even battles. Only experienced traders could return safely.} \\
		Question & \textit{The Silk Road became less important because \underline{\hbox to 8mm{}}}. \\
		Answer & \textit{A.it was made up of different routes} 
		\qquad \qquad	\quad \textit{B.silk trading became less popular} \\
		& \textbf{\textit{C.sea travel provided easier routes}} 
		\qquad \qquad \textit{D.people needed fewer foreign goods}  \\
		\midrule
		Span Extraction &  from SQuAD \cite{rajpurkar2016squad} \\
		\midrule
		Context &  \textit{Robotics is an interdisciplinary branch of engineering and science that includes mechanical engineering, electrical engineering, computer science, and others. Robotics deals with the design, construction, operation, and use of robots, as well as computer systems for their control, sensory feedback, and information processing. These technologies are used to develop machines that can substitute for humans. Robots can be used in any situation and for any purpose, but today many are used in dangerous environments (including bomb detection and de-activation), manufacturing processes, or where humans cannot survive. Robots can take on any form, but some are made to resemble humans in appearance. This is said to help in the acceptance of a robot in certain replicative behaviors usually performed by people. Such robots attempt to replicate walking, lifting, speech, cognition, and basically anything a human can do.} \\
		Question & \textit{What do robots that resemble humans attempt to do?} \\
		Answer & \textit{replicate walking, lifting, speech, cognition}\\
		\midrule
		Free-form &  from DROP \cite{dua2019drop} \\
		\midrule
		Context &  \textit{The Miami Dolphins came off of a 0-3 start and tried to rebound against the Buffalo Bills. After a scoreless first quarter the Dolphins rallied quick with a 23-yard interception return for a touchdown by rookie Vontae Davis and a 1-yard touchdown run by Ronnie Brown along with a 33-yard field goal by Dan Carpenter making the halftime score 17-3. Miami would continue with a Chad Henne touchdown pass to Brian Hartline and a 1-yard touchdown run by Ricky Williams. Trent Edwards would hit Josh Reed for a 3-yard touchdown but Miami ended the game with a 1-yard touchdown run by Ronnie Brown. The Dolphins won the game 38-10 as the team improved to 1-3. Chad Henne made his first NFL start and threw for 115 yards and a touchdown.} \\
		Question & \textit{How many more points did the Dolphins score compare to the Bills by the game's end?} \\
		Answer & \textit{28}\\
		
	\end{tabular}
	
	\label{tab:forms}
\end{table}

\subsection{Task Formulation}
Given the reference document or passage, as the standard form, MRC requires the machine to answer questions about it. The formation of MRC can be described as a tuple $<P, Q, A>$, where $P$ is a passage (context), and $Q$ is a query over the contents of $P$, in which $A$ is the answer or candidate option.

In the exploration of MRC, constructing a high-quality, large-scale dataset is as important as optimizing the model structure. 
Following \citet{chen2018neural},\footnote{We made slight modifications to adapt to the latest emerging types.} the existing MRC variations can be roughly divided into four categories, 1) \textit{cloze-style}; 2) \textit{multi-choice}; 3) \textit{span extraction}, and 4) \textit{free-form prediction}.

\subsection{Typical Datasets}
\paragraph{Cloze-style} For cloze-style MRC, the question contains a placeholder and the machine must decide which word or entity is the most suitable option. The standard datasets are CNN/Daily Mail \cite{hermann2015teaching}, Children's Book Test dataset (CBT) \cite{hill2015goldilocks}, BookTest \cite{bajgar2016embracing}, Who did What \cite{onishi2016who}, ROCStories \cite{mostafazadeh2016corpus}, CliCR \cite{suster2018clicr}.

\paragraph{Multi-choice} This type of MRC requires the machine to find the only correct option in the given candidate choices based on the given passage. The major datasets are MCTest \cite{richardson2013mctest}, QA4MRE \cite{sutcliffe2013QA4MRE}, RACE \cite{lai2017race}, ARC \cite{clark2018think}, SWAG \cite{zellers2018swag}, DREAM \cite{sun2019dream}, etc.

\paragraph{Span Extraction} The answers in this category of MRC are spans extracted from the given passage texts. The typical benchmark datasets are SQuAD \cite{rajpurkar2016squad}, TrivialQA \cite{Joshi2017TriviaQA}, SQuAD 2.0 (extractive with unanswerable questions) \cite{Rajpurkar2018Know}, NewsQA \cite{trischler2017newsqa}, SearchQA \cite{dunn2017searchqa}, etc. 

\paragraph{Free-form Prediction} The answers in this type are abstractive free-form based on the understanding of the passage. The forms are diverse, including generated text spans, yes/no judgment, counting, and enumeration. For free-form QA, the widely-used datasets are MS MACRO \cite{Nguyen2016MSMA}, NarrativeQA \cite{kovcisky2018narrativeqa}, Dureader \cite{he2018dureader}. This category also includes recent conversational MRC, such as CoQA \cite{reddy2019coqa} and QuAC \cite{choi2018quac}, and discrete reasoning types involving counting and arithmetic expression as those in DROP \cite{dua2019drop}, etc.

Except for the variety of formats, the datasets also differ from 1) context styles, e.g., single paragraph, multiple paragraphs, long document, and conversation history; 2) question types, e.g., open natural question, cloze-style fill-in-blank, and search queries; 3) answer forms, e.g., entity, phrase, choice, and free-form texts; 4) domains, e.g., Wikipedia articles, news, examinations, clinical, movie scripts, and scientific texts; 5) specific skill objectives, e.g., unanswerable question verification, multi-turn conversation, multi-hop reasoning, mathematical prediction, commonsense reasoning, coreference resolution. A detailed comparison of the existing dataset is listed in Appendix \S\ref{appendix:datasets}.

\subsection{Evaluation Metrics}
For cloze-style and multi-choice MRC, the common evaluation metric is accuracy. For span-based QA, the widely-used metrics are Exact match (EM) and (Macro-averaged) F1 score. EM  measures the ratio of predictions that match any one of the ground truth answers exactly. F1  score measures the average overlap between the prediction and ground truth answers. For non-extractive forms, such as generative QA, answers are not limited to the original context, so ROUGE-L \cite{lin2004looking} and BLEU \cite{papineni2002bleu} are also further adopted for evaluation.

\subsection{Towards Prosperous MRC}
Most recent MRC test evaluations are based on an online server, which requires to submit the model to assess the performance on the hidden test sets. Official leaderboards are also available for easy comparison of submissions. A typical example is SQuAD. \footnote{\url{https://rajpurkar.github.io/SQuAD-explorer/}.} Open and easy following stimulate the prosperity of MRC studies, which can provide a great precedent for other NLP tasks. We think the success of the MRC task can be summarized as follows:

\begin{itemize}
\item \textbf{Computable Definition}: due to the vagueness and complexity of natural language, on the one hand, a clear and computable definition is essential (e.g., cloze-style, multi-choice, span-based, etc.);
\item \textbf{Convincing Benchmarks}: to promote the progress of any application, technology, open, and comparable assessments are indispensable, including convincing evaluation metrics (e.g., EM and F1), and evaluation platforms (e.g., leaderboards, automatic online evaluations).
\end{itemize}

The definition of a task is closely related to the automatic evaluation. Without computable definitions, there will be no credible evaluation.

\subsection{Related Surveys}
Previous survey papers \cite{zhang2019machine,qiu2019survey,liu2019neural} mainly outlined the existing corpus and models for MRC. Our survey differs from previous surveys in several aspects:

\begin{itemize}
\item Our work goes much deeper to provide a comprehensive and comparative review with an in-depth explanation over the origin and the development of MRC in the broader view of the NLP scenario, paying special focus on the role of CLMs. We conclude that MRC boosts the progress from language processing to understanding, and the theme of MRC is gradually moving from shallow text matching to cognitive reasoning.
\item For the technique side, we propose new taxonomies of the architecture of MRC, by formulating MRC systems as two-stage architecture motivated by cognition psychology and provide a comprehensive discussion of technical methods. We summarize the technical methods and highlights in different stages of MRC development. We show that the rapid improvement of MRC systems greatly benefits from the progress of CLMs.
\item Besides a wide coverage of topics in MRC researches through investigating typical models and trends from MRC leaderboards, our own empirical analysis is also provided. A variety of newly emerged topics, e.g., interpretation of models and datasets, decomposition of prerequisite skills, complex reasoning, low-resource MRC, etc., are also discussed in depth. According to our experience, we demonstrate our observations and suggestions for the MRC researches.\footnote{We are among the pioneers to research neural machine reading comprehension. We pioneered the research direction of employing linguistic knowledge for building MRC models, including morphological segmentation \cite{zhang2018mrc,Zhang2018subword,zhang2018effective}, semantics injection \cite{zhang2019explicit,zhang2020SemBERT}, syntactic guidance \cite{zhang2019sg}, and commonsense \cite{li2020multichoice}. Besides the encoder representation, we investigated the decoder part to strengthen the comprehension, including interactive matching \cite{dcmn20,zhu2020dual}, answer verification \cite{zhang2020retrospective}, and semantic reasoning \cite{zhang2020semantics}. Our researches cover the main topics of MRC. The approaches enable effective and interpretable solutions for real-world applications, such as question answering \cite{zhang2018gaokao}, dialogue and interactive systems \cite{zhang2018dua,zhu2018lingke,zhang2019open}. We also won various first places in major MRC shared tasks and leaderboards, including CMRC-2017, SQuAD 2.0, RACE, SNLI, and DREAM.} 
\end{itemize}

We believe that this survey would help the audience more deeply understand the development and highlights of MRC, as well as the relationship between MRC and the broader NLP community.

\section{Technical Methods}\label{sec:techniques}
\subsection{Two-stage Solving Architecture}
Inspired by dual process theory of cognition psychology \cite{wason1974dual,evans1984heuristic,evans2003two,kahneman2011thinking,evans2017dual,ding2019cognitive}, the cognitive process of human brains potentially involves two distinct types of procedures: contextualized perception (\textit{reading}) and analytic cognition (\textit{comprehension}), where the former gather information in an implicit process, then the latter conduct the controlled reasoning and execute goals. 
Based on the above theoretical basis, in the view of architecture design, a standard reading system (reader) which solves MRC problem generally consists of two modules or building steps: 

1) building a CLM as Encoder; 

2) designing ingenious mechanisms as Decoder according to task characteristics.

\begin{figure}[htb]
%     \centering
    \includegraphics[width=1.0\textwidth]{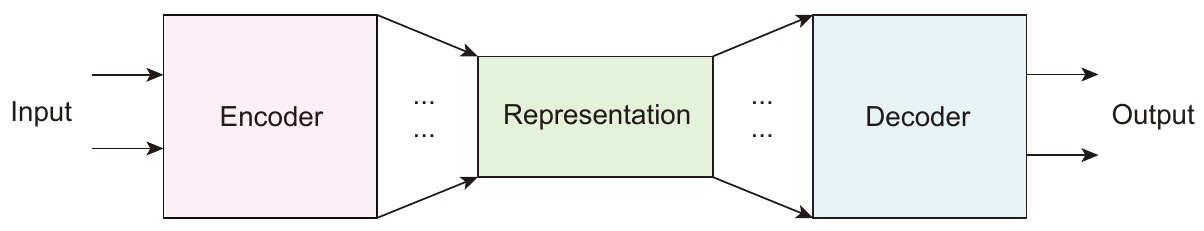}
    \caption{Encoder-Decoder Solving Architecture.}
    \label{fig:enc-dec}
\end{figure}

We find that the generic architecture of MRC system can thus be minimized as the formulation as two-stage solving architecture in the perspective of Encoder-Decoder architecture \cite{sutskever2014sequence}.\footnote{We find that most NLP systems can be formed as such architecture.} General Encoder is to encoder the inputs as contextualized vectors, and Decoder is specific to the detailed task. Figure \ref{fig:enc-dec} shows the architecture.

\subsection{Typical MRC Architecture}

Here we introduce two typical MRC architectures following the above Encoder-Decoder framework, 1) traditional RNN-based \textit{BiDAF} and 2) CLM-powered \textit{BERT}.
\subsubsection{Traditional RNN-based BiDAF}
Before the invention of CLMs, early studies widely adopted RNNs as feature encoders for sequences, among which GRU was the most popular due to the fast and effective performance. The input parts, e.g., passage and question, are fed to the encoder separately. Then, the encoded sequences are passed to attention layers for matching interaction between passage and questions before predicting the answers. The typical MRC model is BiDAF, which is composed of four main layers: 1) encoding layer that transforms texts into a joint representation of the word and character embeddings; 2) contextual encoding that employs BiGRUs to obtain contextualized sentence-level representation;\footnote{Note that BiDAF has the completely contextualized encoding module. Except for the specific module implementation, the major difference with CLMs is that the BIDAF encoder is not pre-trained.}; 3) attention layer to model the semantic interactions between passage and question; 4) answer prediction layer to produce the answer. The first two layers are the counterpart of Encoder, and the last two layers serve the role of Decoder.

\subsubsection{Pre-trained CLMs for Fine-tuning}
When using CLMs, the input passage and question are concatenated as a long sequence to feed CLMs, which merges the encoding and interaction process in RNN-based MRC models. 
Therefore, the general encoder has been well formalized as CLMs, appended with a simple task-specific linear layer as Decoder to predict the answer.

\subsection{Encoder}
The encoder part plays the role of vectorizing the natural language texts into latent space and further models the contextualized features of the whole sequence.

\subsubsection{Multiple Granularity Features} 
\paragraph{Language Units} Utilizing fine-grained features of words was one of the hot topics in previous studies. To solve the out-of-vocabulary (OOV) problem, character-level embedding was once a common unit besides word embeddings \cite{Seo2016Bidirectional,Yang2016Words,dhingra2017gated,zhang2018effective,zhang2019open}. However, character is not the natural minimum linguistic unit, which makes it quite valuable to explore the potential unit (subword) between character and word to model sub-word morphologies or lexical semantics. To take advantage of both word-level and character representations, subword-level representations for MRC were also investigated \cite{zhang2018mrc,Zhang2018subword}. In \citet{zhang2018mrc}, we propose BPE-based subword segmentation to alleviate OOV issues, and further adopt a frequency-based filtering method to strengthen the training of low-frequency words. Due to the highly flexible grained representation between character and word, subword as a basic and effective language modeling unit has been widely used for recent dominant models \cite{devlin2018bert}.

\paragraph{Salient Features}
Linguistic features, such as part-of-speech (POS) and named entity (NE) tags, are widely used for enriching the word embedding \cite{liu2018stochastic}. Some semantic features like semantic role labeling (SRL) tags and syntactic structures also show effectiveness for language understanding tasks like MRC \cite{zhang2020SemBERT,zhang2019sg}. Besides, the indicator feature, like the binary Exact Match (EM) feature is also simple and effective indications, which measures whether a context word is in the question \cite{chen2019convolutional}.

\subsubsection{Structured Knowledge Injection}

Incorporating human knowledge into neural models is one of the primary research interests of artificial intelligence. Recent Transformer-based deep contextual language representation models have been widely used for learning universal language representations from large amounts of unlabeled data, achieving dominant results in a series of NLU benchmarks \cite{peters2018deep,radford2018improving,devlin2018bert,yang2019xlnet,liu2019roberta,lan2019albert}. However, they only learn from plain context-sensitive features such as character or word embeddings, with little consideration of explicit hierarchical structures that exhibited in human languages, which can provide rich dependency hints for language representation. Recent studies show that modeling structured knowledge has shown beneficial for language encoding, which can be categorized into \textit{Linguistic Knowledge} and \textit{Commonsense}. 

\paragraph{Linguistic Knowledge}
Language linguistics is the product of human intelligence, comprehensive modeling of syntax, semantics, and grammar is essential to provide effective structured information for effective language modeling and understanding \cite{zhang2020SemBERT,zhang2019sg,zhang2019explicit,zhou2019limit}.
\paragraph{Commonsense}
 At present, reading comprehension is still based on shallow segment extraction, semantic matching in limited text, and lack of modeling representation of commonsense knowledge. Human beings have learned commonsense through the accumulation of knowledge over many years. In the eyes of human beings, it is straightforward that ``the sun rises in the east and sets in the west", but it is challenging to learn by machine. Commonsense tasks and datasets were proposed to facilitate the research, such as ROCStories \cite{mostafazadeh2016corpus}, SWAG \cite{zellers2018swag}, CommonsenseQA \cite{talmor2019commonsenseqa}, ReCoRD \cite{zhang2018record}, and Cosmos QA \cite{huang2019cosmos}. Several commonsense knowledge graphs are available as the prior knowledge sources, including ConceptNet \cite{speer2017conceptnet}, WebChild \cite{tandon2017webchild} and ATOMIC \cite{sap2019atomic}. It is an important research topic to let machines learn and understand human commonsense effectively to be used in induction, reasoning, planning, and prediction.

\subsubsection{Contextualized Sentence Representation} 

Previously, RNNs, such as LSTM, and GRU were seen as the best choice
in sequence modeling or language models. However, the recurrent architectures have a fatal flaw, which is hard to parallel in the training process, limiting the computational efficiency. 
\citet{NIPS2017_7181} proposed Transformer, based entirely
on self-attention rather than RNN or Convolution. Transformer can not only achieve parallel calculations but
also capture the semantic correlation of any span. Therefore,
more and more language models tend to choose it to be the
feature extractor. Pre-trained on a large-scale textual corpus, these CLMs well serve as the powerful encoders for capturing contextualized sentence representation.

\subsection{Decoder}
After encoding the input sequences, the decoder part is used for solving the task with the contextualized sequence representation, which is specific to the detailed task requirements. For example, the decoder is required to select a proper question for multi-choice MRC or predict an answer span for span-based MRC.

\begin{figure*}
    \centering
    \includegraphics[width=1.0\textwidth]{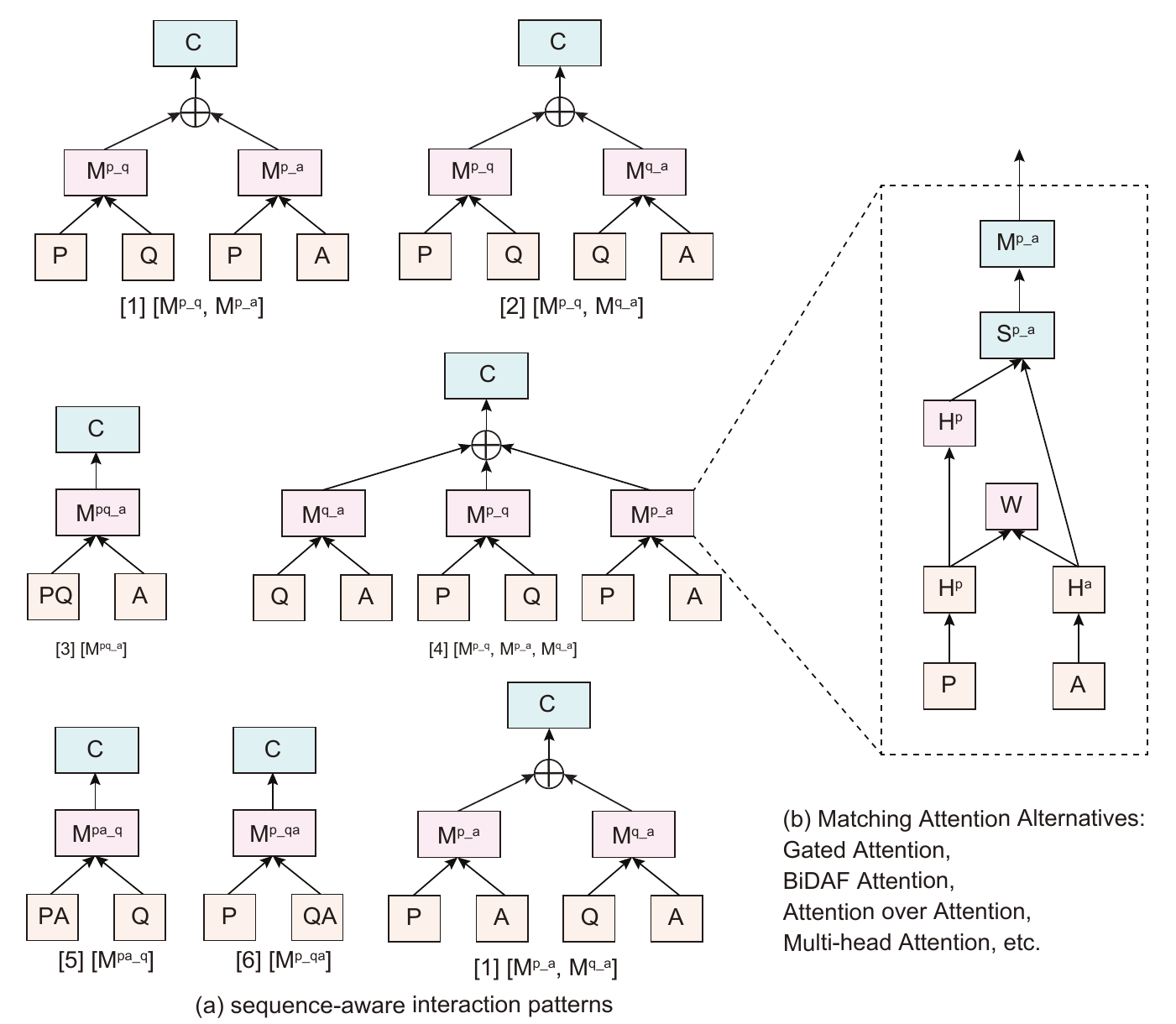}
    \caption{Designs of matching network.}
    \label{fig:matching}
\end{figure*}

Not until recently keep the primary focuses of nearly all MRC systems on the encoder side, i.e., the deep pre-trained models  \cite{devlin2018bert}, as the systems may simply and straightforwardly benefit from a strong enough encoder. Meanwhile, little attention is paid to the decoder side of MRC models \cite{hu2019read,back2020neurquri}, though it has been shown that better decoder or better manner of using encoder still has a significant impact on MRC performance, no matter how strong the encoder (i.e., the adopted pre-trained CLM) it is \cite{dcmn20}. In this part, we discuss the decoder design in three aspects: 1) \textit{matching network}; 2) \textit{answer pointer}, 2) \textit{answer verifier}, and 3) \textit{answer type predictor}. 

\subsubsection{Matching Network} 
The early trend is a variety of attention-based interactions between passage and question, including: Attention Sum \cite{kadlec2016text}, Gated Attention \cite{dhingra2017gated}, Self-matching  \cite{Wang2017Gated}, BiDAF Attention  \cite{Seo2016Bidirectional}, Attention over Attention \cite{Cui2017Attention}, and Co-match Attention \cite{wang2018co}.

Some work is also investigating the attention-based interactions of passage and question in the era of Transformer-based backbones, such as dual co-match attention \cite{dcmn20,zhu2020dual}. 
Figure \ref{fig:matching} presents the exhaustive patterns of matching considering three possible sequences: passage ($P$), question ($Q$), and answer candidate option ($A$).\footnote{Though many well-known matching methods only involve passage and question as for cloze-style and span-based MRC, we present a more general demonstration by also considering multi-choice types that have three types of input, and the former types are also included as counterparts.} The sequences, $P$, $Q$ or $A$, can be concatenated together as one, for example, $PQ$ denotes the concatenation of $P$ and $Q$. $M$ is defined as the matching operation. For example, $M^{p\_a}$ models the matching between the hidden states of $P$ and $A$. We depict the simple but widely-used matching attention $M$ in Figure \ref{fig:matching}-(b) for example, whose formulation is further described in \S\ref{sec:interaction} for detailed reference. However, the study of the matching mechanisms has come to a bottleneck facing the already powerful CLM encoders, which are essentially interactive to model paired sequences.

\subsubsection{Answer Pointer}
Span prediction is one of the major focuses of MRC tasks. Most models predict the answer by generating the start position and the end position corresponding to the estimated answer span. Pointer network \cite{vinyals2015pointer} was used in early MRC models \cite{wang2016machine,Wang2017Gated}.

For training the model to predict the answer span for an MRC task, standard maximum-likelihood method is used for predicting exactly-matched (EM) start and end positions for an answer span. It is a strict objective that encourages exact answers at the cost of penalizing nearby or overlapping answers that are sometimes equally accurate. To alleviate the issue and predict more acceptable answers, reinforcement learning algorithm based self-critical policy learning was adopted to measure the reward as word overlap between the predicted answer and the ground truth, so as to optimize towards the F1 metric instead of EM metric for span-based MRC \cite{xiong2018dcn,hu2018reinforced}.

\begin{figure*}
    \centering
    \includegraphics[width=1.0\textwidth]{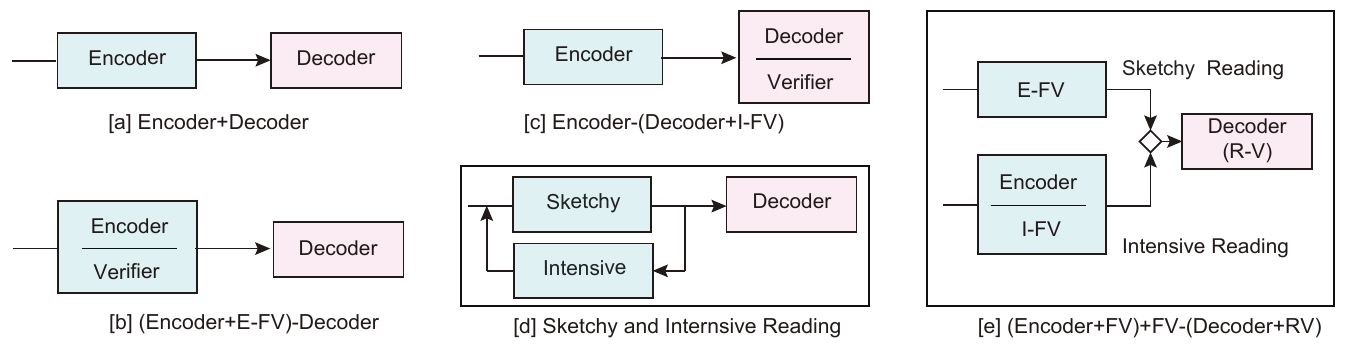}
    \caption{Designs of answer verifier.}
    \label{fig:verifier}
\end{figure*}

% \paragraph{Question classifier}
\subsubsection{Answer Verifier}
For the concerned MRC challenge with unanswerable questions, a reader has to handle two aspects carefully: 1) give the accurate answers for answerable questions; 2) effectively distinguish the unanswerable questions, and then refuse to answer. Such requirements complicate the reader's design by introducing an extra verifier module or answer-verification mechanism. Figure \ref{fig:verifier} shows the possible designs of the verifiers. The variants are mainly three folds (the formulations are elaborated in \S\ref{sec:assess}):

1) Threshold-based answerable verification (TAV). The verification mechanism can be simplified an answerable threshold over predicted span probability that is broadly used by powerful enough CLMs for quickly building readers \cite{devlin2018bert,zhang2020SemBERT}. 

2) Multitask-style verification (Intensive). Mostly, for module design, the answer span prediction and answer verification are trained jointly with multitask learning (Figure \ref{fig:verifier}(c)). 
\citet{liu2018stochastic} appended an empty word token to the context and added a simple classification layer to the reader. 
\citet{hu2019read} used two types of auxiliary loss, independent span loss to predict plausible answers and independent no-answer loss to decide the answerability of the question. Further, an extra verifier is adopted to decide whether the predicted answer is entailed by the input snippets (Figure \ref{fig:verifier}(b)). 
\citet{back2020neurquri} developed an attention-based satisfaction score to compare question embeddings with the candidate answer embeddings. It allows explaining why a question is classified as unanswerable by showing unmet conditions within the question (Figure \ref{fig:verifier}(c)). 
\citet{zhang2019sg} proposed a linear verifier layer to context embedding weighted by start and end distribution over the context words representations concatenated to special pooled \texttt{[CLS]} token representation for BERT (Figure \ref{fig:verifier}(c)). 

3) External parallel verification (Sketchy). \citet{zhang2020retrospective} proposed a Retro-Reader that integrates two stages of reading and verification strategies: 1) sketchy reading that briefly touches the relationship of passage and question, and yields an initial judgment; 2) intensive reading that verifies the answer and gives the final prediction (Figure \ref{fig:verifier}(d)).
In the implementation, the model is structured as a rear verification (RV) method that combines the multitask-style verification as internal verification (IV), and external verification (EV) from a parallel module trained only for answerability decision, which is both simple and practicable with basically the same performance, which results in a parallel reading module design at last as the model shown in Figure \ref{fig:verifier}(e).

\subsubsection{Answer Type Predictor}
Most of the neural reading models \cite{Seo2016Bidirectional,Wang2017Gated,yu2018qanet} are usually designed to extract a continuous span of text as the answer. For more open and realistic scenarios, where answers are involved with various types, such as numbers, dates, or text strings, several pre-defined modules are used to handle different kinds of answers \cite{dua2019drop,gupta2019neural,hu2019multi}.

\begin{table}[t]
\caption{Loss functions for MRC. CE: categorical crossentropy, BCE: binary crossentropy, MSE: mean squared error.}
\begin{tabular}{lcccp{5.6cm}}
\toprule
Type & CE & BCE &  MSE  &\\
\midrule
Cloze-style & \cmark & & \\
Span-based & \cmark & & \\
\quad + (binary) verification & \cmark & \cmark & \cmark \\
\quad + yes/no & \cmark & \cmark & \cmark \\
\quad + count & \cmark & & \\
Multi-choice & \cmark & & \\
\end{tabular}
\label{tab:task-loss}
\end{table}

\subsection{Training Objectives}
Table \ref{tab:task-loss} shows the training objectives for different types of MRC. The widely-used objective function is cross-entropy. For some specific types, such as binary answer verification, categorical crossentropy, binary crossentropy, and mean squared error are also investigated \cite{zhang2020retrospective}. Similarly, for tasks involve yes or no answers, the three alternative functions are also available. For counting, previous researches tend to model it as multi-class classification task using crossentropy \cite{dua2019drop,hu2019multi,ran2019numnet}.

\begin{table*}
%     \centering
\scriptsize
             \caption{Typical MRC models for comparison of Encoders on SQuAD 1.1 leaderboard. TRFM is short for Transformer. Although MRC models often employ ensembles for better performance, the results are based single models to avoid extra influence in ensemble models. * QANet and BERT used back translation and TriviaQA  dataset \cite{Joshi2017TriviaQA} for further data augmentation, respectively. The improvements $\uparrow$ are calculated based on the result (\textit{italic}) on Match-LSTM.}
  \small
  \def\arraystretch{1.2}
  \setlength{\tabcolsep}{5pt}
    {
    \begin{tabular}{llccrr}
            \toprule
%             \multirow{2}{*}{\textbf{Models}} & 
%             \multirow{2}{*}{\textbf{Encoder}} & \multirow{2}{*}{\textbf{Augmentation}} & \multicolumn{2}{c}{\textbf{SQuAD1.1}} & \\
            \textbf{Models}  & \textbf{Encoder}
             &  \textbf{EM} & \textbf{F1} &  $\uparrow$ \textbf{EM} & $\uparrow$ \textbf{F1}\\
\midrule
Human \cite{Rajpurkar2018Know}  & -  & 82.304	& 91.221 & - & -\\ 
            \hdashline
            Match-LSTM \cite{wang2016machine} & RNN &  \textit{64.744} &    \textit{73.743} & - & -  \\
            DCN \cite{xiong2016dynamic}  &RNN &  66.233 & 75.896  & 1.489 & 2.153 \\
            Bi-DAF \cite{Seo2016Bidirectional} & RNN  & 67.974 &    77.323 & 3.230 & 3.580 \\
        
            Mnemonic Reader \cite{DBLP:journals/corr/HuPQ17} &RNN &  70.995    & 80.146 & 6.251 & 6.403  \\

            Document Reader \cite{chen2017reading} & RNN  & 70.733 & 79.353 & 5.989 & 5.610 \\
        
            DCN+ \cite{xiong2017dcn+}  & RNN & 75.087 & 83.081  & 10.343 & 9.338 \\
                
            r-net \cite{Wang2017Gated}   & RNN  & 76.461 & 84.265 & 11.717 & 10.522 \\
            
            MEMEN \cite{pan2017memen}  & RNN & 78.234 & 85.344  & 13.490 & 11.601 \\
    
            QANet \cite{yu2018qanet}*   &TRFM & 80.929 & 87.773 & 16.185 & 14.030  \\
            
            \hdashline    
            \multicolumn{6}{c}{\textit{CLMs}}\\
            ELMo \cite{peters2018deep}  & RNN  & 78.580 & 85.833 & 13.836 & 12.090 \\

            BERT \cite{devlin2018bert}* & TRFM  & 85.083 & 91.835 & 20.339 & 18.092 \\
    
            SpanBERT \cite{joshi2020spanbert} & TRFM  & 88.839 & 94.635 & 24.095 & 20.892 \\
        
            XLNet \cite{yang2019xlnet}   & TRFM-XL  & 89.898 & 95.080 & 25.154 & 21.337 \\
    
    \end{tabular}
    }
    \label{mrcmodels1}
\end{table*}

\begin{table*}
%     \centering
\scriptsize
             \caption{Typical MRC models for comparison of Encoders on SQuAD 2.0 and RACE leaderboard. TRFM is short for Transformer. The D-values $\uparrow$ are calculated based on the results (\textit{italic}) on BERT for SQuAD 2.0 and GTP$_{v1}$ for RACE.}
  \small
  \def\arraystretch{1.2}
  \setlength{\tabcolsep}{5pt}
    {
    \begin{tabular}{llcccc}
            \toprule
            \textbf{Models}  & \textbf{Encoder}
             & \textbf{SQuAD 2.0} & $\uparrow$ \textbf{F1} & \textbf{RACE}  &  $\uparrow$ \textbf{Acc} \\
\midrule
Human \cite{Rajpurkar2018Know}  & - & 91.221 & - & -\\ 
    GPT$_{v1}$ \cite{radford2018improving} & TRFM &  - &  - & \textit{59.0} & - \\
    BERT \cite{devlin2018bert}  & TRFM  & \textit{83.061} & - & \text{72.0} & -\\
    \quad SemBERT \cite{zhang2020SemBERT}   & TRFM  & 87.864 & 4.803 & - & -\\ 
    
    \quad SG-Net \cite{zhang2019sg} &   TRFM  & 87.926 & 4.865 &  - & -  \\
    
    \quad RoBERTa \cite{liu2019roberta}  & TRFM  & 89.795 &  6.734 & 83.2 & 24.2 \\ 

    \quad ALBERT \cite{lan2019albert}   & TRFM  & 90.902 &  7.841 & 86.5 & 27.5 \\
    XLNet \cite{yang2019xlnet}  & TRFM-XL  & 90.689 & 7.628 & 81.8 & 22.8 \\
    ELECTRA  \cite{clark2019electra}   & TRFM  & 91.365 & 8.304 & - & - \\
    
    \end{tabular}
    }
    \label{mrcmodels2}
\end{table*}
\section{Technical Highlights}\label{sec:reallyworks}
In this part, we summarize the previous and recent dominant techniques by reviewing the systems for the flagship datasets concerning the main types of MRC, cloze-type CNN$/$DailyMail \cite{hermann2015teaching}, multi-choice RACE \cite{lai2017race}, and span extraction SQuAD \cite{rajpurkar2016squad,Rajpurkar2018Know}. 
Tables \ref{mrcmodels1},\ref{mrcmodels2},\ref{tab:contrib-lms},\ref{tab:dec-match},\ref{tab:cnn} show the statistics, from which we summarize the following observations and thoughts (we will elaborate the details in the subsequent sections):

1) \textbf{CLMs greatly boost the benchmark of current MRC}. Deeper, wider encoders carrying large-scale knowledge become a new major theme. The upper bound of the encoding capacity of deep neural networks has not been reached yet; however, training such CLMs are very time-consuming and computationally expensive. Light and refined CLMs would be more friendly for real-world and common usage, which can be realized by designing more ingenious models and learning strategies \cite{lan2019albert}, as well as knowledge distillation \cite{jiao2019tinybert,sanh2019distilbert}.

2) \textbf{Recent years witness a decline of matching networks}. Early years witnessed a proliferation of attention-based mechanisms to improve the interaction and matching information between passage and questions, which work well with RNN encoders. After the popularity of CLMs, the advantage disappeared. Intuitively, the reason might be that CLMs are interaction-based models (e.g., taking paired sequences as input to model the interactions), but not good feature extractors. This difference might be the pre-training genre of CLMs, and also potentially due to the transformer architecture.
It is also inspiring that it promotes a transformation from shallow text matching into a more complex knowledge understanding of MRC researches to some extent.

3) \textbf{Besides the encoding sides, optimizing the decoder modules is also essential for more accurate answers}. Especially for SQuAD2.0 that requires the model to decide if a question is answerable, training a separate verifier or multitasking with verification loss generally works.\footnote{We notice that jointly multitasking verification loss and answer span loss has been integrated as a standard module in the released codes in XLNet and ELECTRA for SQuAD2.0.}

4) \textbf{Data augmentation from similar MRC datasets sometimes works}. Besides some work reported using TraiviaQA \cite{Joshi2017TriviaQA} or NewsQA \cite{Joshi2017TriviaQA} datasets as extra training data, there were also many submissions whose names contain terms about data augmentation. Similarly, when it comes to the CLMs realm, there is rarely work that uses augmentation. Besides, the pre-training of CLMs can also be regarded as data augmentation, which is highly potential for the performance gains.

In the following part, we will elaborate on the major highlights of the previous work. We also conduct a series of empirical studies to assess simple tactic optimizations as a reference for interested readers (\S\ref{sec:assess}).

\subsection{Reading Strategy}
Insights on the solutions to MRC challenges can be drawn from the cognitive process of humans. Therefore, some interesting reading strategies are proposed based on human reading patterns, such as Learning to Skim Text \cite{yu2017learning}, learning to stop reading \cite{shen2017reasonet}, and our proposed retrospective reading \cite{zhang2020retrospective}. Also, \cite{sun2019improving} proposed three general strategies: back and forth reading, highlighting, and self-assessment to improve non-extractive MRC.

\begin{table}
\caption{The contributions of CLMs. * indicates results that depend on additional external training data. $\dagger$ indicate the result is from \citet{yang2019xlnet} as it was not reported in the original paper \cite{devlin2018bert}. Since the final results were reported by the largest models, we listed the $large$ models for XLNet, BERT, RoBERTa, ELECTRA, and $xxlarge$ model for ALBERT. GPT is reported as the v1 version. }
  \setlength{\tabcolsep}{5.2pt}
\begin{tabular}{llllllllll}
\toprule
\multirow{2}{*}{Method}  & \multirow{2}{*}{Tokens}  &  \multirow{2}{*}{Size}  &\multirow{2}{*}{Params} & \multicolumn{2}{c}{SQuAD1.1} & \multicolumn{2}{c}{SQuAD2.0} & \multirow{2}{*}{RACE} & \\
 &  &  &   & Dev & Test & Dev & Test & \\
\midrule
ELMo & 800M & - & 93.6M  & 85.6 & 85.8 & - & - & - \\
GPT$_{v1}$  & 985M & - & 85M  & - & - & - & - &   59.0\\
XLNet$_{large}$   & 33B  & - & 360M & 94.5 & 95.1* & 88.8 & 89.1* &  81.8 \\
BERT$_{large}$ & 3.3B & 13GB & 340M  & 91.1 & 91.8* & 81.9 & 83.0 & 72.0$\dagger$ \\
% RoBERTa & Wikipedia + BooksCorpus & 16GB  &  355M & 93.6 & 87.3  & \\
\quad RoBERTa$_{large}$  & - & 160GB &  355M & 94.6  & - & 89.4 & 89.8 & 83.2\\
\quad ALBERT$_{xxlarge}$  & - &157GB & 235M  & 94.8 & - & 90.2 & 90.9 & 86.5\\
ELECTRA$_{large}$  &  33B & - & 335M  & 94.9 & - & 90.6 & 91.4 & - \\
\end{tabular}
\label{tab:contrib-lms}
\end{table}

\begin{figure}[htb]
     \centering
    \includegraphics[width=0.8\textwidth]{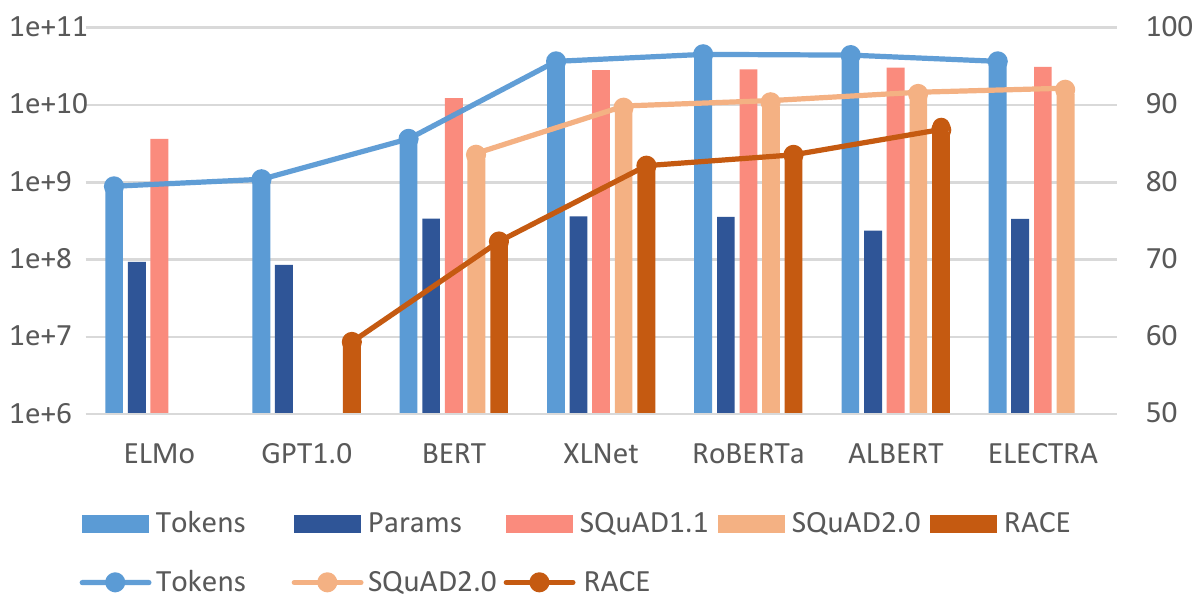}
    \caption{The contribution of the sizes of pre-trained corpus and CLMs. The right axis is the main metric for the statistics. The numbers of tokens and parameters are normalized by $\log_{10^6}(x)+50$ where $x$ denotes the original number. The left axis corresponds to the original values of tokens and parameters for easy reference.}
    \label{fig:params}
\end{figure}

\subsection{CLMs Become Dominant}
As shown in Table \ref{tab:contrib-lms}, CLMs improve the MRC benchmarks to a much higher stage. Besides the contextualized sentence-level representation, the advance of CLMs is also related to the much larger model size and large-scale pre-training corpus. From Table \ref{tab:contrib-lms} and the further illustration in Figure \ref{fig:params}, we see that both the model sizes and the scale of training data are increasing remarkably, that contribute the downstream MRC model performance.\footnote{The influence of model parameters can also be easily verified at the SNLI leaderboard: \url{https://nlp.stanford.edu/projects/snli/}.}

\subsection{Data Augmentation}
Since most high-quality MRC datasets are human-annotated and inevitably relatively small, another simple method to boost performance is data augmentation. Early effective data augmentation is to inject extra similar MRC data for training a specific model. Recently, using CLMs, which pre-trained on large-scale unlabeled corpora, can be regarded as a kind of data augmentation as well.
\paragraph{Training Data Augmentation} There are various methods to provide extra data to train a more powerful MRC model, including: 1) Combining various MRC datasets as training data augmentation (TDA) \cite{DBLP:journals/corr/abs-1904-06652,yang2019data}; 2) Multi-tasking \cite{xu2018multi,fisch2019mrqa}; 3) Automatic question generation, such as back translation \cite{yu2018qanet} and synthetic generation \cite{du2017learning,du2017identifying,kim2019improving,zhu2019learning,alberti2019synthetic}.
However, we find the gains become small when using CLMs, which might already contain the most common and important knowledge between different datasets.
\paragraph{Large-scale Pre-training} Recent studies showed that CLMs well acquired linguistic information through pre-training \cite{clark2019does,ettinger2020bert} (more discussions in Section \S\ref{sec:Interpretability}), which is potential to the impressive results on MRC tasks. 

\begin{table}[t]
\caption{Typical MRC models for comparisons of decoding designs on multi-choice RACE test sets. The matching patterns correspond to those notations in Figure \ref{fig:matching}. M: RACE-M, H: RACE-H. M, H, RACE are the accuracy on two subsets and the overall test sets, respectively.}
  \setlength{\tabcolsep}{2.8pt}
\begin{tabular}{llccc}
\toprule
Model & Matching & M & H &  RACE \\
\midrule
\multicolumn{2}{l}{Human Ceiling Performance \cite{lai2017race}} & 95.4 & 94.2 & 94.5 \\
\multicolumn{2}{l}{Amazon Mechanical Turker \cite{lai2017race}} & 85.1 & 69.4 & 73.3 \\
\midrule
HAF \cite{zhu2018hierarchical} & $[M^{P\_A}; M^{P\_Q}; M^{Q\_A}]$ &45.0 & 46.4 &46.0\\
MRU  \cite{tay2018multi} & $[M^{P\_Q\_A}]$  &57.7 & 47.4 &50.4\\
HCM \cite{wang2018co} & $[M^{P\_Q}; M^{P\_A}]$ & 55.8 & 48.2 &50.4\\
MMN \cite{tang2019multi}& $[M^{Q\_A}; M^{A\_Q}; M^{P\_Q}; M^{P\_A}]$ &61.1 & 52.2 &54.7\\
GPT \cite{radford2018improving} &$[M^{P\_Q\_A}]$  &62.9 & 57.4 &59.0\\
RSM \cite{sun2019improving}& $[M^{P\_QA}]$ &69.2 & 61.5 &63.8\\
DCMN \cite{zhang2019dual} & $[M^{PQ\_A}]$ & 77.6 & 70.1 & 72.3 \\
OCN \cite{ran2019option} &$[M^{P\_Q\_A}]$  &76.7 & 69.6 &71.7 \\
BERT$_{large}$ \cite{pan2019improving}& $[M^{P\_Q\_A}]$ & 76.6 & 70.1 & 72.0 \\
XLNet \cite{yang2019xlnet}& $[M^{P\_Q\_A}]$ &85.5 & 80.2 &81.8 \\
\quad + DCMN+ \cite{dcmn20} &$[M^{P\_Q}; M^{P\_O}; M^{Q\_O}]$ & 86.5 & 81.3 & 82.8 \\
RoBERTa \cite{liu2019roberta}& $[M^{P\_Q\_A}]$  & 86.5 & 81.8 & 83.2 \\
\quad + MMM \cite{jin2019mmm} &$[M^{P\_Q\_A}]$ &89.1 & 83.3 & 85.0 \\
ALBERT \cite{jin2019mmm} &$[M^{P\_Q\_A}]$ &    89.0 & 85.5 & 86.5 \\
\quad + DUMA \cite{zhu2020dual}& $[M^{P\_QA}; M^{QA\_P}]$ & 90.9 & 86.7 & 88.0\\
Megatron-BERT \cite{shoeybi2019megatron}& $[M^{P\_Q\_A}]$ & 91.8 & 88.6 & 89.5\\
\end{tabular}
\label{tab:dec-match}
\end{table}

\begin{table}[htb]
    \caption{ Results on cloze CNN$/$DailyMail test sets. UA: unidirectional attention. BA: bidirectional attention. The statistics are from \citet{Seo2016Bidirectional}.}
% \parbox{.45\linewidth}{
 \setlength{\tabcolsep}{5pt}
    \centering
    \begin{tabular}{lccccc}
       \toprule
       \multirow{2}{*}{Method} & \multirow{2}{*}{Att. Type} & \multicolumn{2}{c}{CNN} & \multicolumn{2}{c}{DailyMail}\\
         & & val & test & val & test\\
        \midrule
        Attentive Reader~\cite{hermann2015teaching} &  UA & 61.6 & 63.0 & 70.5 & 69.0 \\
        AS Reader~\cite{kadlec2016text} & UA & 68.6 & 69.5 & 75.0 & 73.9 \\
        Iterative Attention~\cite{sordoni2016iterative} & UA & 72.6 & 73.3 & - & - \\
        Stanford AR~\cite{chen2016thorough} & UA & 73.8 & 73.6 & 77.6 & 76.6 \\
        GAReader~\cite{dhingra2017gated} & UA & 73.0 & 73.8 & 76.7 & 75.7 \\
        AoA Reader~\cite{Cui2017Attention} & BA & 73.1 & 74.4 & - & - \\
        BiDAF \cite{Seo2016Bidirectional} & BA &  76.3  &  76.9  & 80.3  & 79.6 \\
    \end{tabular}

    \label{tab:cnn}
%}
\end{table}

\subsection{Decline of Matching Attention}
As the results shown in Tables \ref{tab:dec-match}-\ref{tab:cnn}, it is easy to notice that the attention mechanism is the
key component in previous RNN-based MRC systems.\footnote{We roughly summarize the matching methods in the previous work using our model notations, which meet their general ideas except some calculation details.}

We see that bidirectional attention (BA) works better than unidirectional one, and co-attention is a superior matching method, which indicate the advance of more rounds of matching that would be effective at capturing more fine-grained information intuitively.
When using CLMs as the encoder, we observe that the explicit passage and question attention could only show quite marginal, or even degradation of performance. The reason might be that CLMs are interaction-based matching models \cite{qiao2019understanding} when taking the whole concatenated sequences of passage and question. It is not suggested to be employed as a representative model. \citet{bao2019inspecting} also reported similar observations, showing that the unified modeling of sequences in BERT outperforms previous networks that separately treat encoding and matching.

After contextualized encoder by the CLMs, the major connections for reading comprehension might have been well modeled, and the vital information is aggregated to the representations of special tokens, such as \texttt{[CLS]} and \texttt{[SEP]} for BERT. We find that the above encoding process of CLMs is quite different from that in traditional RNNs, where the hidden states of each token are passed successively in one direction, without mass aggregation and degradation of representations.\footnote{Although the last hidden state is usually used for the overall representation, the other states may not suffer from degradation like in multi-head attention-based deep CLMs.} The phenomenon may explain why interactive attentions between input sequences work well with RNN-based feature extractors but show no obvious advantage in the realm of CLMs.

\subsection{Tactic Optimization}
\paragraph{The objective of answer verification}
For answer verification, modeling the objective as classification or regression would have a slight influence on the final results. However, the advance might vary based on the backbone network, as some work took the regression loss due to the better performance \cite{yang2019xlnet}, while the recent work reported that the classification would be better in some cases \cite{zhang2020retrospective}.
\paragraph{The dependency inside answer span}
Recent CLM-based models simplified the span prediction part as independent classification objectives. However, the end position is related to the start predictions. As a common method in early works \cite{Seo2016Bidirectional}, jointly integrating the start logits and the sequence hidden states to obtain the end logits is potential for further enhancement. Another neglected aspect recently is the dependence of all the tokens inside an answer span, instead of considering only the start and end positions. 
\paragraph{Re-ranking of candidate answers}
Answer reranking is adapted to mimic the process of double-checking. A simple strategy is to use N-best reranking strategy after generating answers from neural networks \cite{Cui2017Attention,wang2018evidence,wang2018multi,wang2018joint,hu2019retrieve}. Unlike previous work that ranks candidate answers, \citet{hu2019multi} proposed an arithmetic expression reranking mechanism to rank expression candidates that are decoded by beam search, to incorporate their context information during reranking to confirm the prediction further.

\subsection{Empirical Analysis of Decoders}\label{sec:assess}
To gain insights on how to further improve
MRC, we report our attempts to improve model performance with general and straightforward tactic optimizations for the widely-used SQuAD2.0 dataset that does not rely on the backbone model. The methods include three types, \textit{Verification}, \textit{Interaction}, and \textit{Answer Dependency}.\footnote{In this part, we intend to intuitively show what kinds of tactic optimizations potentially work, so we brief the details of the methods and report the best results as a reference after hyper-parameter searching. We recommend interested readers to read our technical report \cite{zhang2020retrospective} for the details of answer verification and sequence interactions. Our sources are publicly available at \url{https://github.com/cooelf/AwesomeMRC}.}

\subsubsection{Baseline}
We adopt BERT$_{large}$ \cite{devlin2018bert} and ALBERT$_{xxlarge}$ \cite{lan2019albert} as our baselines. 

\paragraph{Encoding} The input sentence is first tokenized to word pieces (subword tokens). Let $T=\{t_1,\dots,t_L\}$ denote a sequence of subword tokens of length $L$. For each token, the input embedding is the sum of its token embedding, position embedding, and token-type embedding. Let $X = \{x_1, \dots, x_L\}$ be the outputs of the encoder, which are embedding features of encoding sentence words of length $L$. The input embeddings are then fed into the deep Transformer \cite{NIPS2017_7181} layers for learning contextual representations. 
Let $X^{g} = \{x^{g}_{L}, \dots, x^{g}_{L} \}$ be the features of the $g$-th layer. The features of the $g+1$-th layer, $x^{g+1}$ is computed by
\begin{align}
\tilde{h}_{i}^{g+1} &= \sum_{m=1}^{M}W_{m}^{g+1}\left \{ \sum_{j=1}^{n}A_{i,j}^{m}\cdot V_{m}^{g+1}x_{j}^{g} \right \},\label{mha} \\
h_{i}^{g+1} &= \textit{LayerNorm}(x_{i}^{g}+\tilde{h}_{i}^{g+1}), \\
\tilde{x}_{i}^{g+1} &= W_{2}^{g+1}\cdot \textit{GELU}(W_{1}^{g+1}h_{i}^{g+1}+ b_{1}^{g+1}) + b_{2}^{g+1}, \label{ffn}\\
x_{i}^{g+1} &= \textit{LayerNorm}(h_{i}^{g+1} + \tilde{x}_{i}^{g+1}),
\end{align}
where $m$ %in Eq. (\ref{mha}) 
is the index of the attention heads, and $A_{i,j}^{m}\propto \mathrm{exp}[(Q_{m}^{g+1}x_{i}^{g})^{\top }(K_{m}^{g+1}x_{j}^{g})]$ denotes the attention weights between elements $i$ and $j$ in the $m$-th head, which is normalized by $\sum_{j=1}^{N}A_{i,j}^{m}=1$. $W_{m}^{g+1}, Q_{m}^{g+1}, K_{m}^{g+1}$ and $V_{m}^{g+1}$ are learnable weights for the $m$-th attention head, $W_{1}^{g+1}, W_{2}^{g+1}$ and $b_{1}^{g+1},b_{2}^{g+1}$ %in Eq. (\ref{ffn})
are learnable weights and biases, respectively. Finally, we have  last-layer hidden states of the input sequence $\textbf{H} = \{h_1, \dots, h_L\}$ as the contextualized representation of the input the sequence.

\paragraph{Decoding}The aim of span-based MRC is to find a span in the passage as answer, thus we employ a linear layer with SoftMax operation and feed $\textbf{H}$ as the input to obtain the start and end probabilities, $s$ and $e$:
    \begin{equation}
    s, e \propto \textit{SoftMax}(\textit{Linear}(\textbf{H})).
    \end{equation}
    
\paragraph{Threshold based answerable verification (TAV)}\label{TAV} For unanswerable question prediction, given output start and end probabilities $s$ and $e$, and the verification probability $v$, we calculate the has-answer score $score_{has}$ and the no-answer score $score_{na}$:

    \begin{equation}
    \begin{split}
    score_{has} & =\max (s_{k_1} + e_{k_2}),1 < k_1 \le k_2 \le L, \\
    score_{na} &= s_1+e_1,
    \end{split}
    \end{equation}
    where $s_1$ and $e_1$ denote the corresponding logits for the special token \texttt{[CLS]} as in BERT-based models used for answer verification \cite{devlin2018bert,lan2019albert}.
%     where $\lambda_{1}$ and $ \lambda_{2}$ are weights.
    We obtain a difference score between \textit{has-answer} score and the \textit{no-answer} score as final score. An answerable threshold $\delta$ %is set to determine whether the question is answerable, which 
	is set and determined %computed in linear time with dynamic programming 
	according to the development set. The model predicts the answer span that gives the \textit{has-answer} score if the final score is above the threshold $\delta$, and null string otherwise.
    
\paragraph{Training Objective}The training objective of answer span prediction is defined as cross entropy loss for the start and end predictions, 
    \begin{equation}
    \mathbb{L}^{span} = -\frac{1}{N}\sum_{i}^{N}[\log (p_{y^{s}_{i}}^s)+\log (p_{y^{e}_{i}}^e)],
    \end{equation}
    where $y^{s}_{i}$ and $y^{e}_{i}$ are respectively ground-truth start and end positions of example $i$. $N$ is the number of examples.

\subsubsection{Verification} Answer verification is vital for MRC tasks that involve unanswerable answers. We tried to add an external separate classifier model that is the same as the MRC model except for the training objective (E-FV). We weighted the predicted verification logits and original heuristic no-answer logits to decide whether the question is answerable. Besides, we also investigated adding multitasking the original span loss with verification loss as an internal front verifier (I-FV). The internal verification loss can be a cross-entropy loss (I-FV-CE), binary cross-entropy loss (I-FV-BE), or regression-style mean square error loss (I-FV-MSE).

The pooled first token (special symbol, \texttt{[CLS]}) representation $h_1 \in \textbf{H}$, as the overall representation of the sequence,\footnote{Following the initial practice of BERT-tyle models, the first token (special symbol, \texttt{[CLS]}) representation is supposed to be the overall representation of the sequence owing to the pre-training objective.} is passed to a fully connection layer to get classification logits or regression score. Let $\hat{y}_{i}  \propto \textit{Linear}(h_1)$ denote the prediction and $y_{i}$ is the answerability target, the three alternative loss functions are as defined as follows:

(1) For cross entropy as loss function for the classification verification: 
\begin{equation}
\begin{split}
    \hat{y}_{i,k} &= \textit{SoftMax}(\textit{Linear}(h_1)),\\
    \mathbb{L}^{ans} &= -\frac{1}{N}\sum_{i=1}^{N}\sum_{k=1}^{K}\left [ y_{i,k}\log\hat{y}_{i,k} \right ],
    \end{split}
\end{equation}
where $K$ is the number of classes. In this work, $K=2$.

(2) For binary cross entropy as loss function for the classification verification: 
\begin{equation}
\begin{split}
    \hat{y}_{i} &= \textit{Sigmoid}(\textit{Linear}(h_1)),\\
    \mathbb{L}^{ans} &= -\frac{1}{N}\sum_{i=1}^{N}\left [ y_{i}\log\hat{y}_{i} + (1-y _{i})\log(1-\hat{y}_{i})\right ].
 \end{split}
\end{equation}

(3) For the regression verification, %the loss function is the 
mean square error is adopted as its loss function: 
\begin{equation}
\begin{split}
    \hat{y}_{i} &= \textit{Linear}(h_1),\\
\mathbb{L}^{ans} &= \frac{1}{N}\sum_{i=1}^{N}(y_{i}-\hat{y}_{i})^{2}.
\end{split}
\end{equation}

During training, the joint loss function for FV is the weighted sum of the span loss and verification loss.
\begin{equation}
\mathbb{L} = \alpha_{1}\mathbb{L}^{span} + \alpha_{2} \mathbb{L}^{ans},
\end{equation}
where $\alpha_{1}$ and $\alpha_{2}$ are weights. We set $\alpha_{1}=\alpha_{2}=0.5$ for our experiments.

We empirically find that training with joint loss can yield better results, so we also report the results of 1) summation of all the I-FV losses (All I-FVs: I-FV-CE, I-FV-BE, and I-FV-MSE), 2) combination of external and internal verification (All I-FVs + E-FV) by calculating the sum (denoted as $v$) of the logits of E-FV and I-FVs as the final answerable logits. In the later scenario, the TAV is rewritten as,
\begin{equation}
\begin{split}
score_{has} & =\max (s_{k_1} + e_{k_2}),1 < k_1 \le k_2 \le L, \\
score_{na} &= \lambda_{1} (s_1+e_1) +  \lambda_{2} v,
\end{split}
\end{equation}
where $\lambda_{1}$ and $ \lambda_{2}$ are weights. We set $\lambda_{1}=\lambda_{2}=0.5$.
    
\subsubsection{Interaction}\label{sec:interaction}
To obtain the representation of each passage and question, we split the last-layer hidden state $\textbf{H}$ into $\textbf{H}^Q$ and $\textbf{H}^P$ as the representations of the question and passage, according to its position information. Both of the sequences are padded to the maximum length in a minibatch. Then, we investigate two potential question-aware matching mechanisms, 1) Transformer-style multi-head attention (MH-ATT), and 2) traditional dot attention (DT-ATT).

$\bullet$ \textbf{Multi-head Attention} 
    We feed the $\textbf{H}^Q$ and $\textbf{H}$ to a revised one-layer multi-head attention layer inspired by \citet{lu2019vilbert}.\footnote{We do not use $\textbf{H}^P$ because $\textbf{H}$ achieved better results in our preliminary experiments.} Since the setting is $\textbf{Q}=\textbf{K}=\textbf{V}$ in multi-head self attention,\footnote{In this work, $Q,K,V$ correspond to the items $Q_{m}^{g+1}x_{i}^{g}, K_{m}^{g+1}x_{j}^{g})$ and $V_{m}^{g+1}x_{j}^{g}$, respectively.} which are all derived from the input sequence, we replace the input to $\textbf{Q}$ with $\textbf{H}$, and both of $\textbf{K}$ and $\textbf{V}$ with $\textbf{H}^Q$ to obtain the question-aware context representation $\textbf{H}'$.
    
    $\bullet$ \textbf{Dot Attention}
    Another alternative is to feed $\textbf{H}^Q$ and $\textbf{H}$ to a traditional matching attention layer \cite{Wang2017Gated}, by taking the question presentation $\textbf{H}^Q$ as the attention to the %context 
    representation $\textbf{H}^{C}$:
    \begin{equation}
    \begin{split}
    \textbf{M} &= \textit{SoftMax}(\textbf{H}(\textbf{W}_p\textbf{H}^Q+\textbf{b}_{p}\otimes\textbf{e}_{q})^\mathsf{T}),\\
    \textbf{H}' &= \textbf{M}\textbf{H}^Q,
    \end{split}
    \label{eq2:Image_Representation}
    \end{equation}
    where $\textbf{W}_q$ and $\textbf{b}_{q}$ are learnable parameters. $\textbf{e}_{q}$ is a all-ones vector and used to repeat the bias vector into the matrix. $\textbf{M}$ denotes the weights assigned to the different hidden states in the concerned two sequences. $\textbf{H}'$ is the weighted sum of all the hidden states and it represents how the vectors in $\textbf{H}$ can be aligned to each hidden state in $\textbf{H}^Q$.
    
    Finally, the %contextual 
    representation $\textbf{H}'$ is used for the later predictions as described in the \textit{decoding} and \textit{TAV} section above.

\subsubsection{Answer Dependency}
Recent studies separately use $\textbf{H}$ to predict the start and end spans for the answer, neglecting the dependency of the start and end representations. We model the  dependency between start and end logits by concatenating the start logits and $\textbf{H}$ through a linear layer to obtain the end logits:
\begin{equation}
e = \textit{Linear}([s; \textbf{H}]),
\end{equation}
where $[;]$ denotes concatenation.

\begin{table}
    \caption{\label{table-assess} Results (\%) of different decoder mechanisms on the SQuAD2.0 dev set. Part of the numbers of \textit{verification} and \textit{interactions} are adapted from our previous work \cite{zhang2020retrospective} (slight update with further hyperparameter tuning).}
    {
                \begin{tabular}{lllllp{5.2cm}}
                    \toprule
                    \multirow{2}{*}{Method}  & \multicolumn{2}{c}{BERT}& \multicolumn{2}{c}{ALBERT} \\
                    & \textbf{EM} & \textbf{F1} & \textbf{EM} & \textbf{F1} &\\
                    \midrule
                    \textit{Baseline}  & 78.8 & 81.7 &  87.0 & 90.2 \\
                                
                   \multicolumn{5}{l}{\textit{Interaction}} \\
                    \quad     + MH-ATT &  78.8 & 81.7  & 87.3 & 90.3 \\
                    \quad     + DT-ATT  & 78.3  & 81.4  & 86.8 & 90.0 \\
                
                \multicolumn{5}{l}{\textit{Verification}}\\
                    \quad     + E-FV & 79.1 & 82.1 & 87.4 & 90.6 \\
                    \quad     + I-FV-CE  & 78.6 & 82.0 & 87.2 & 90.3 \\
                    \quad     + I-FV-BE  & 78.8 & 81.8 & 87.2 & 90.2  \\
                    \quad     + I-FV-MSE & 78.5 & 81.7 & 87.3 & 90.4 \\
                %     \cdashline{1-5}
                    \quad     + All I-FVs  & 79.4 & 82.1 & 87.5  & 90.6  \\
                    \quad     + All I-FVs + E-FV  & 79.8 & 82.7  & 87.7  & 90.8  \\
                    
                    \multicolumn{5}{l}{\textit{Answer Dependency}} \\
                    \quad     + SED & 79.1 & 81.9 & 87.3 & 90.3 \\
                %     \quad     + SLD & 78.6 & 81.6 & 87.3 & 90.3 \\
            
                \end{tabular}
            }
\end{table}

\subsubsection{Findings}

Table \ref{table-assess} shows the results. Our observations are as follows:

\begin{itemize}
\item For answer verification, either of the front verifiers boosts the baselines, and integrating all the verifiers can yield even better results.
\item Adding extra interaction layers after the strong CLMs could only yield marginal improvement, which verifies the CLMs' strong ability to capture the relationships between passage and question. 
\item Answer dependency can effectively improve the exact match score, which can intuitively help yield a more exactly matched answer span.
\end{itemize}

\section{Trends and Discussions}\label{sec:trends}
\subsection{Interpretability of Human-parity Performance} \label{sec:Interpretability}
Recent years witnessed frequent reports of super human-parity results in MRC leaderboards, which further stimulated the research interests of investigating what the `real' ability of MRC systems, and what kind of knowledge or reading comprehension skills the systems have grasped. The interpretation appeal to aspects of CLM models, MRC datasets, and models.
\paragraph{For CLM models} Since CLM models serve as the basic module for contextualized text representation, fingering out what the knowledge captured, especially what linguistic capacities CLMs process confer upon models, is critical for fine-tuning downstream tasks, so is for MRC. There are heated discussions about what CLM models learn recently. Recent work has tried to give the explanation by investigating the attention maps from the multi-head attention layers \cite{clark2019does}, and conducting diagnostic tests \cite{ettinger2020bert}. \citet{clark2019does} found that attention heads correspond well to linguistic notions of syntax and coreference. \citet{ettinger2020bert}
introduced a suite of diagnostic tests to assess the linguistic competencies of BERT, indicating that BERT performs sensitivity to role reversal and same-category distinctions. Still, it struggles with challenging inferences and role-based event prediction, and it shows obvious failures with the meaning of negation.

\paragraph{For MRC datasets and models} So far, the MRC system is still a black box, and it is very
risky to use it in many scenarios in which we have to know
how and why the answer is obtained. It is critical to deeply investigate the explanation of the MRC models or design an explainable MRC architecture. Although MRC datasets are emerging rapidly and the corresponding models continuously show impressive results, it still hard to interpret what MRC systems learned, so is the benchmark capacity of the diversity of MRC datasets \cite{sugawara2018makes,sugawara2019assessing,schlegel2020framework}. The common arguments are the overestimated ability of MRC systems as MRC models do not necessarily provide human-level understanding, due to the unprecise benchmarking on the existing datasets. Although there are many models show human-parity scores so far, we cannot say that they successfully perform human-level reading comprehension. The issue mainly lies within the low interpretability of both of the explicit internal processing of currently prevalent neural models, and what is measured by the datasets. Many questions can be answered correctly by the model that do not necessarily require grammatical and complex reasoning. For example, \citet{jia2017adversarial} and \citet{wallace2019trick} provided manually crafted adversarial examples to show that MRC systems are easily distracted. \citet{sugawara2019assessing} also indicated that most of the questions already answered correctly by the model do not necessarily require grammatical and complex reasoning. The distractors can not only assess the vulnerability of the current models but also serve as salient hard negative samples to strengthen model training \cite{gao2019generating}.

Besides, as discussed in our previous work \cite{zhang2020retrospective}, since current results are relatively high in various MRC benchmark datasets, with relatively marginal improvement, it is rarely confirmed that produced results are statistically significant than baseline. For the
reproducibility of models, it is necessary to conduct statistical tests in evaluating MRC models.

\subsection{Decomposition of Prerequisite Skills} \label{sec:data_construct}
As the experience of human examinations, good comprehension requires different dimensions of skills. The potential solution for our researches is to decompose the skills required by the dataset and take skill-wise evaluations, thus provide more explainable and convincing benchmarking of model capacity. Further, it would be beneficial to consider following a standardized format, which can make it simpler to conduct cross-dataset evaluations \cite{fisch2019mrqa}, and train a comprehensive model that can work on different datasets with specific skills.

Regarding the corresponding benchmark dataset construction, it is no coincidence that SQuAD datasets turned out a success and have served as the standard benchmark. Besides the high quality and specific focus of the datasets, an online evaluation platform that limits the submission frequency also ensures the convincing assessment. 
On the other hand, it is natural to be cautious for some comprehensive datasets with many complex question types, which requires many solver modules, as well as processing tricks--we should report convincing evaluation with detailed evaluation on separate types or subtasks, instead of just pursuing overall SOTA results. Unless honestly reporting and unifying the standards of these processing tricks, the evaluation would be troublesome and hard to replicate.

\subsection{Complex Reasoning}
Most of the previous advances have focused on \textit{shallow} QA tasks
that can be tackled very effectively by existing retrieval and matching-based techniques. Instead of measuring the comprehension and understanding of the QA systems in question, these tasks test merely the capability of a method to focus attention on specific words and pieces of text. To better align the progress in the field of QA with the expectations that we have of human performance and behavior when solving such tasks, a new class of questions, e.g., ``complex" or ``challenge" reasoning, has been a hot topic. Complex reasoning can most generally be thought of as instances that require intelligent behavior and reasoning on the part of a machine to solve. 

As the knowledge, as well as the questions themselves, become more complex and specialized, the process of understanding and answering these questions comes to resemble human expertise in specialized domains. Current examples of such complex reasoning tasks,
where humans presently rule the roost, include customer support, standardized testing in education, and domain-specific consultancy services, such as medical and legal advice. The study of such complex reasoning would be promising for machine intelligence from current perception to next-stage cognition.

 Recent studies have been proposed for such kind of comprehension, including multi-hop QA \cite{welbl2018constructing,yang2018hotpotqa} and conversational QA \cite{reddy2019coqa,choi2018quac}.
To deal with the complex multi-hop relationship, dedicated mechanism design is needed for multi-hop commonsense reasoning. Besides, structured knowledge provides a wealth of prior commonsense context, which promotes the research on the fusion of multi-hop commonsense knowledge between symbol and semantic space in recent years \cite{lin2019kagnet,ma2019towards}. For conversational QA, modeling multi-turn dependency requires extra memory designs to capture the context information flow and solve the problems precisely and consistently \cite{huang2018flowqa}.

Regarding technical side, graph-based neural networks (GNN), including graph attention network, graph convolutional network, and graph recurrent network have been employed for complex reasoning \cite{song2018exploring,qiu2019dynamically,chen2019graphflow,jiang2019explore,tu2019select,tu2020graph}. The main intuition behind the design of GNN based models is to answer questions that require to explore and reason over multiple scattered pieces of evidence, which is similar to human's interpretable step-by-step problem-solving behavior. 
Another theme appearing frequently in machine learning in general is the revisiting of the existing models and how they perform in a fair experimental setting. \citet{shao2020graph} raised a concern that graph structure may not be necessary for multi-hop reasoning, and graph-attention can be considered as a particular case of self-attention as that used in CLMs. We can already see a transformation from heuristic applications of GNNs to more sound approaches and discussions about the effectiveness of graph models. For future studies, an in-depth analysis of GNNs, as well as the connections and differences between GNNs and CLMs would be inspiring.

\subsection{Large-scale Comprehension} 
Most current MRC systems are based on the hypothesis of given passages as reference context. However, for real-world MRC applications, the reference passages, even documents, are always lengthy and detail-riddled. However, recent LM based models work slowly or even unable to process long texts. The ability of knowledge extraction is especially needed for open-domain and free-form QA whose reference texts are usually large-scale \cite{guu2020realm}. A simple solution is to train a model to select the relevant information pieces by calculating the similarity with the question \cite{chen2017reading,clark2018simple,htut2018training,tan2018s,wang2018multi,zhang2019examination,yan2019deep,min2018efficient,nishida2019multi}. Another technique is to summarize the significant information of the reference context, by taking advantage of text summarization or compression \cite{li2019explicit}.

\subsection{Low-resource MRC}
Low-resource processing is a hot research topic since most of the natural languages lack abundant annotated data \cite{wang2019adversarial,zhang2019neural}. Since most MRC studies are based on the English datasets, there exists a considerable gap for other languages that do not have high-quality MRC datasets. Such a situation can be alleviated by transferring the well-trained English MRC models through domain adaptation \cite{wang2019adversarial}, and training semi-supervised \cite{yang2017semi,zhang2018gaokao} or multilingual MRC systems \cite{liu2019xcmrc,lee2019learning,cui2019cross}.

The other major drawback exposed in MRC systems is the inadequate knowledge transferability \cite{talmor2019multiqa} as they are trained, and even over-fitted on specific datasets. Since most of the famous datasets are built from Wikipedia articles, the apparent benefits from CLMs might be the same or similar text patterns contained in the training corpus, e.g., context, topics, etc.  It remains a significant challenge to design robust MRC models that are immune to real noise. It is also essential to build NLP systems that generalize across domains, especially unseen domains \cite{fisch2019mrqa}.

\subsection{Multimodal Semantic Grounding}
Compared with human learning, the current pure text processing model performance is relatively weak, because this kind of model only learns the text features, without the perception of the external world, such as visual information. In human learning, people usually understand the world through visual images, auditory sounds, words, and other modes. Human brain perceives the world through multimodal semantic understanding. Therefore, multimodal semantic modeling is closer to human perception, which is conducive to a more comprehensive language understanding. It remains an open problem when and how to make full use of different modalities to improve reading comprehension and inference. A related research topic is visual question answering \cite{balanced_vqa_v2}, which aims to answer questions according to a given image. However, it is still in the early stage of research as the QA is concerned with only one image context. As a more practical scenario, jointly modeling diverse modalities will be potential research interests, and beneficial for real-world applications, e.g., E-commerce customer support. For example, given the mixed text, image, and audio background conversation context, the machine is required to give responses to the inquiry accordingly. With the continuous advance of computational power, we believe the joint supervision of auditory, tactile, and visual sensory information together with the language will be crucial for next-stage cognition.

% \subsection{Better inference module design}

\subsection{Deeper But Efficient Network}
Besides the high-quality benchmark datasets, the increase the computational resources, e.g., GPU, enables us to build deeper and wider networks. The last decade witnessed the traditional feature extractor from the RNN to deep transformers, with a larger capacity for contextualized modeling. In the future, we are confident that much deeper and stronger backbone frameworks will be proposed with the rapid development of GPU capacity and further boost the MRC system benchmark performance. In the meantime, smaller and refined systems, potentially through knowledge distillation from large models, also occupy a certain market, which relies on rapid and accurate reading comprehension solving ability for real-world application.

\section{Conclusion}\label{sec:conclu}
This work comprehensively reviews the studies of MRC in the scopes of background, definition, development, influence, datasets, technical and benchmark highlights, trends, and opportunities. We first briefly introduced the history of MRC and the background of contextualized language models. Then, we discussed the role of contextualized language models and the influence of MRC to the NLP community. The previous technical advances were summarized in the framework of Encoder to Decoder. After going through the mechanisms of MRC systems, we showed the highlights in different stages of MRC studies. Finally, we summarized the trends and opportunities. The basic views we have arrived at are that 1) MRC boosts the progress from language processing to understanding; 2) the rapid improvement of MRC systems greatly benefits from the progress of CLMs; 3) the theme of MRC is gradually moving from shallow text matching to cognitive reasoning.

\appendix

\appendixsection{Machine Reading Comprehension Datasets}
\label{appendix:datasets}

This appendix lists existing machine reading comprehension datasets along with their answer styles, dataset size, type of corpus, sourcing methods, and focuses.  Part of the statistics is borrowed from \citet{sugawara2020prerequisites}. 
% These datasets are sorted in chronological order by year of publication. 
\textit{Ans} denotes answer styles where  \textit{Ex} is answer extraction by selecting a span in the given context, and \textit{FF} is free-form answering. \textit{NA} denotes that unanswerable questions are involved, and YN means yes or no answers. \textit{Size} indicates the size of the whole dataset, including training, development, and test sets. \textit{Src} represents how the questions are sourced where \textit{X} means questions written by experts, \textit{C} by crowdworkers, \textit{A} by machines with an automated manner, and \textit{Q} are search-engine queries. Note that the boundary between cloze-style and multi-choice datasets is not clear sometimes; for example, some candidate choices may be provided for cloze tests, such as Story Cloze Test \cite{mostafazadeh2017lsdsem} and CLOTH \cite{xie2018large}. In our taxonomy, we regard the fix-choice tasks whose candidates are in a fixed number as multi-choice. In addition, some datasets are composed of different types of subtasks; we classify them according to the main types with special notations in \textit{Ans} column.

\begin{table*}\footnotesize
  \def\arraystretch{1.5}
  \setlength{\tabcolsep}{6pt}
    \caption{Cloze-style MRC datasets.}
  \begin{tabular}{ccccccccccc} \toprule
Name & Size  & Domain & Src & \linestack{Feature} \\ \midrule 
 \linestack{CNN/ DailyMail \cite{hermann2015teaching}}  & 1.4M & \linestack{news  article} & A & \linestack{entity cloze} \\ 
 \linestack{Children's Book Test \cite{hill2015goldilocks}} & 688K & narrative & A & \linestack{large-scale automated} \\ 
\linestack{BookTest \cite{bajgar2016embracing}} & 14.1M & narrative & A & \linestack{similar to CBT, \\but much larger} \\ 
\linestack{Who did What \cite{onishi2016who}}  & 200K & \linestack{news  article} & A & \linestack{cloze of person name} \\ 
\linestack{ROCStories \cite{mostafazadeh2016corpus}}  & 50K*5 & \linestack{narrative} & C & \linestack{Commonsense Stories} \\ 
\linestack{CliCR \cite{suster2018clicr}} & 100K & \linestack{clinical case \\ text} & A & \linestack{cloze style queries} \\ 
\end{tabular}
\end{table*}

\begin{table*}[htb]\footnotesize
  \def\arraystretch{1.5}
  \setlength{\tabcolsep}{6.5pt}
    \caption{Multi-choice MRC datasets.}
  \begin{tabular}{cccccccccccc} \toprule
Name & Size & Domain & Src & \linestack{Feature} \\ \midrule 
 \linestack{QA4MRE \cite{sutcliffe2013QA4MRE}} & 240 & \linestack{technical \\ document} & X & \linestack{exam-level questions} \\ 
  \linestack{MCTest\\\cite{richardson2013mctest}} & 2.6K & \linestack{written \\ story} & C & \linestack{children-level narrative} \\ 

    \linestack{RACE \cite{lai2017race}}  & 100K & \linestack{language \\ exam} & X & \linestack{middle/high school \\ English exam in China} \\ 
 \linestack{Story Cloze Test\\\cite{mostafazadeh2017lsdsem}}  & 3.7K & \linestack{written \\ story} & C & \linestack{98,159 stories for training} \\
  \linestack{TextbookQA\\\cite{kembhavi2017smarter}} & 26K & textbook & X & \linestack{figures involved} \\ 
  \linestack{ARCT\\\cite{habernal2018argument}}  & 2.0K & \linestack{debate \\ article} & C/X & \linestack{reasoning on argument} \\ 
    \linestack{CLOTH\\\cite{xie2018large}} & 99K & various & X & \linestack{cloze exam} \\ 
    \linestack{MCScript\\\cite{ostermann2018mcscript}}  & 30K & \linestack{written \\ story} & C & \linestack{commonsense reasnoing, \\ script knowledge} \\ 
    \linestack{ARC\\\cite{clark2018think}} & 8K & \linestack{science \\ exam} & X & \linestack{retrieved documents \\ from textbooks} \\ 
     \linestack{MultiRC\\\cite{khashabi2018looking}} & 6K & \linestack{various \\ documents} & C & \linestack{multi-sentence reasoning} \\ 
      \linestack{SWAG\\\cite{zellers2018swag}} & 113K & \linestack{video \\ captions} & M & \linestack{commonsense reasoning} \\ 
      \linestack{OpenbookQA\\\cite{mihaylov2018suit}} & 6.0K & textbook & C & \linestack{commonsense reasoning} \\ 
 \linestack{RecipeQA\\\cite{yagcioglu2018recipeqa}} & 36K & \linestack{recipe \\ script} & A & \linestack{multimodal questions} \\ 
  \linestack{Commonsense QA\\\cite{talmor2019commonsenseqa}} & 12K & ConceptNet & C & \linestack{commonsense reasoning} \\ 
  \linestack{DREAM\\\cite{sun2019dream}} & 10K & \linestack{language \\ exam} & X & \linestack{dialogue-based, \\ 6.4k multi-party dialogues} \\ 
    \linestack{MSCript 2.0\\\cite{ostermann2019mcscript2}} & 20K & narrative & C & \linestack{commonsense reasoning, \\ script knowledge} \\ 
 \linestack{HellaSWAG\\\cite{zellers2019hellaswag}} & 70K & \linestack{web \\ snippet} & A & \linestack{commonsense reasoning, \\ WikiHow and ActivityNet} \\ 
  \linestack{CosmosQA\\\cite{huang2019cosmos}} & 36K & narrative & C & \linestack{commonsense reasoning} \\ 
  \linestack{QuAIL\\\cite{rogers2020getting}} & 15K & various & C & \linestack{prerequisite real tasks} \\
\end{tabular}
\end{table*}

\begin{table*}\footnotesize
  \def\arraystretch{1.5}
  \setlength{\tabcolsep}{2.5pt}
    \caption{Span-extraction MRC datasets.}
  \begin{tabular}{ccccccccccccc} \toprule
Name & Ans & Size & Domain & Src & \linestack{Feature} \\ \midrule 
  \linestack{SQuAD 1.1\\\cite{rajpurkar2016squad}} & Ex & 100K & Wikipedia & C & \linestack{large-scale crowdsourced} \\ 

 \linestack{NewsQA\\\cite{trischler2017newsqa}} & Ex & 120K & \linestack{news \\ article} & C & \linestack{blindly created questions} \\ 
 \linestack{SearchQA\\\cite{dunn2017searchqa}} & Ex & 140K & \linestack{web \\ snippet} & C/X & \linestack{snippets from search engine} \\ 
  \linestack{TriviaQA\\\cite{Joshi2017TriviaQA}} & Ex & 650K & \linestack{web \\ snippet} & C/X & \linestack{trivia questions} \\ 
 \linestack{Quasar\\\cite{dhingra2017quasar}} & Ex & 80K & \linestack{web \\ snippet} & Q & \linestack{search queries} \\ 
  \linestack{AddSent SQuAD\\\cite{jia2017adversarial}} & Ex & 3.6K & Wikipedia & C & \linestack{distracting sentences injected} \\
  \linestack{QAngaroo\\\cite{welbl2018constructing}} & Ex & 50K & \linestack{Wikipedia, \\ MEDLINE} & A & \linestack{multi-hop reasoning} \\ 

    \linestack{DuoRC\\\cite{saha2018duorc}} & Ex & 186K & \linestack{movie \\ script} & C & \linestack{commonsense reasoning, \\ multi-sentence reasoning} \\ 
 \linestack{ProPara\\\cite{dalvi2018tracking}} & Ex & 2K & \linestack{science \\ exam} & A & \linestack{procedural understanding} \\ 

 \linestack{Multi-party Dialog\\\cite{ma2018challenging}} & Ex & 13K & \linestack{TV show \\ transcript} & A & \linestack{1.7k crowdsourced dialogues, \\ cloze query} \\ 
 \linestack{SQuAD 2.0\\\cite{Rajpurkar2018Know}} & Ex (+NA) & 100K & Wikipedia & C & \linestack{unanswerable questions} \\

 \linestack{Textworlds QA\\\cite{labutov2018multi}} & Ex & 1.2M & \linestack{generated \\ text} & A & \linestack{simulated worlds, \\ logical reasoning} \\ 
  \linestack{emrQA\\\cite{pampari2018emrqa}} & Ex & 400K & \linestack{clinical \\ documents} & A & \linestack{using annotated logical forms \\ on i2b2 dataset} \\ 
 \linestack{HotpotQA\\\cite{yang2018hotpotqa}} & Ex (+YN) & 113K & Wikipedia & C & \linestack{multi-hop reasoning} \\ 
  \linestack{ReCoRD\\\cite{zhang2018record}} & Ex & 120K & \linestack{news \\ article} & C & \linestack{commonsense reasoning, \\ cloze query} \\

    \linestack{Natural Questions\\\cite{kwiatkowski2019natural}} & Ex (+YN) & 323K & Wikipedia & Q/C & \linestack{short/long answer styles} \\

  \linestack{Quoref\\\cite{dasigi2019quoref}} & Ex & 24K & Wikipedia & C & \linestack{coreference resolution} \\ 

     \linestack{TechQA \\\cite{castelli2019techqa}} & Ex (+NA) & 1.4K & IT support & X & \linestack{technical support domain, \\ domain-adaptation} \\
\end{tabular}
\end{table*}

\begin{table*}\footnotesize
  \def\arraystretch{1.5}
  \setlength{\tabcolsep}{2.5pt}
    \caption{Free-form MRC datasets.}
  \begin{tabular}{ccccccccccccc} \toprule
Name & Ans & Size & Domain & Src & \linestack{Feature} \\ \midrule 
 \linestack{bAbI\\\cite{weston2015bAbI}} & FF & \linestack{10K * \\ 20} & \linestack{generated \\ text} & A & \linestack{prerequisite toy tasks} \\
  \linestack{LAMBADA\\\cite{paperno2016lambada}} & FF & 10K & narrative & C & \linestack{hard language modeling} \\ 
 \linestack{WikiReading\\\cite{hewlett2016wikireading}} & FF & 18M & Wikipedia & A & \linestack{super large-scale dataset} \\ 
  \linestack{MS MARCO\\\cite{Nguyen2016MSMA}} & FF & 100K & \linestack{web \\ snippet} & Q & \linestack{description on web snippets} \\ 
  \linestack{NarrativeQA\\\cite{kovcisky2018narrativeqa}} & FF & 45K & \linestack{movie \\ script} & C & \linestack{summary/full story tasks} \\ 
   \linestack{DuReader\\\cite{he2018dureader}} & FF & 200K & \linestack{web \\ snippet} & Q/C & \linestack{Chinese, \\ Baidu Search/Knows} \\ 
    \linestack{QuAC\\\cite{choi2018quac}} & FF (+YN) & 100K & Wikipedia & C & \linestack{dialogue-based, \\ 14k dialogs} \\ 
     \linestack{ShARC\\\cite{saeidi2018interpretation}} & YN$^*$ & 32K & \linestack{web \\ snippet} & C & \linestack{reasoning on rules taken from \\ government documents} \\ 
      \linestack{CoQA\\\cite{reddy2019coqa}} & FF (+YN) & 127K & Wikipedia & C & \linestack{dialogue-based, \\ 8k dialogs} \\ 
        \linestack{BoolQ\\\cite{clark2019boolq}} & YN & 16K & Wikipedia & Q/C & \linestack{boolean questions, \\ subset of Natural Questions} \\
            \linestack{PubMedQA\\\cite{jin2019pubmedqa}} & YN & 273.5K & PubMed & X/A & \linestack{biomedical domain, \\ 1k expert questions} \\
           \linestack{DROP\\\cite{dua2019drop}} & FF & 96K & Wikipedia & C & \linestack{discrete reasoning} \\ 
\end{tabular}
\end{table*}

\starttwocolumn
\bibliography{compling_style}

\end{document}